\documentclass[10pt,twocolumn,letterpaper]{article}
\usepackage{cvpr}      

\makeatletter
\@namedef{ver@everyshi.sty}{}
\makeatother

\usepackage{graphicx}
\usepackage{amsmath}
\usepackage{amssymb}
\usepackage{booktabs}

\usepackage{pifont}
\usepackage[euler]{textgreek}
\usepackage{calrsfs}
\usepackage{svg}
\DeclareMathAlphabet{\pazocal}{OMS}{zplm}{m}{n}
\usepackage{ctable} 
\usepackage{array, makecell} %
\usepackage{tabularx,booktabs}
\usepackage{multirow}
\usepackage{amssymb} 
\usepackage{pifont}%
\usepackage{float} 
\newcommand{\expnumber}[2]{{#1}\mathrm{e}{#2}}

\usepackage[pagebackref,breaklinks,colorlinks]{hyperref}

\usepackage[capitalize]{cleveref}
\crefname{section}{Sec.}{Secs.}
\Crefname{section}{Section}{Sections}
\Crefname{table}{Table}{Tables}
\crefname{table}{Tab.}{Tabs.}

\def\cvprPaperID{47} 
\def\confName{CVPR}
\def\confYear{2022}

\begin{document}

\title{GenISP: Neural ISP for Low-Light Machine Cognition}

\author{Igor Morawski$^1$ Yu-An Chen$^1$ Yu-Sheng Lin$^1$ Shusil Dangi$^3$ Kai He$^3$ Winston H. Hsu$^{1,2}$ 
\\
\\
$^1$National Taiwan University \qquad $^2$Mobile Drive Technology \qquad $^3$Qualcomm Inc. 
}

\maketitle

\begin{abstract}
Object detection in low-light conditions remains a challenging but important problem with many practical implications. Some recent works show that, in low-light conditions, object detectors using raw image data are more robust than detectors using image data processed by a traditional ISP pipeline. To improve detection performance in low-light conditions, one can fine-tune the detector to use raw image data or use a dedicated low-light neural pipeline trained with paired low- and normal-light data to restore and enhance the image. However, different camera sensors have different spectral sensitivity and learning-based models using raw images process data in the sensor-specific color space. Thus, once trained, they do not guarantee generalization to other camera sensors. We propose to improve generalization to unseen camera sensors by implementing a minimal neural ISP pipeline for machine cognition, named GenISP, that explicitly incorporates Color Space Transformation to a device-independent color space. We also propose a two-stage color processing implemented by two image-to-parameter modules that take down-sized image as input and regress global color correction parameters. Moreover, we propose to train our proposed GenISP under the guidance of a pre-trained object detector and avoid making assumptions about perceptual quality of the image, but rather optimize the image representation for machine cognition. At the inference stage, GenISP can be paired with any object detector. We perform extensive experiments to compare our proposed method to other low-light image restoration and enhancement methods in an extrinsic task-based evaluation and validate that GenISP can generalize to unseen sensors and object detectors. Finally, we contribute a low-light dataset of 7K raw images annotated with 46K bounding boxes for task-based benchmarking of future low-light image restoration and low-light object detection.
\end{abstract}
\vspace{-4em}


\newcommand\blfootnote[1]{%
  \begingroup
  \renewcommand\thefootnote{}\footnote{#1}%
  \addtocounter{footnote}{-1}%
  \endgroup
}

\blfootnote{Dataset: \href{https://github.com/igor-morawski/RAW-NOD}{https://github.com/igor-morawski/RAW-NOD}}

\section{Introduction}
\label{sec:intro}
\begin{figure}[t!]
\begin{center}

\includegraphics[width=38mm]{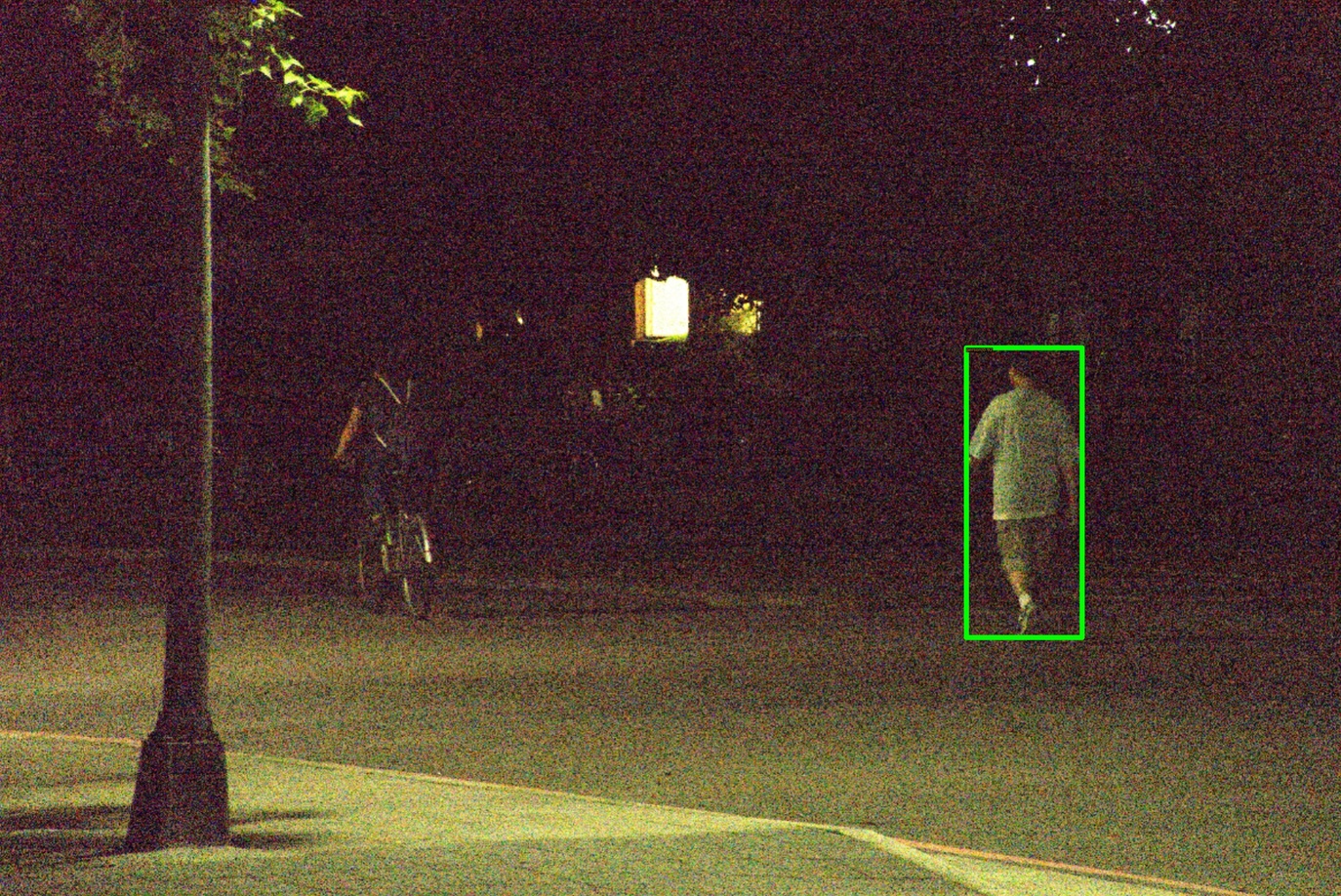}
\includegraphics[width=38mm]{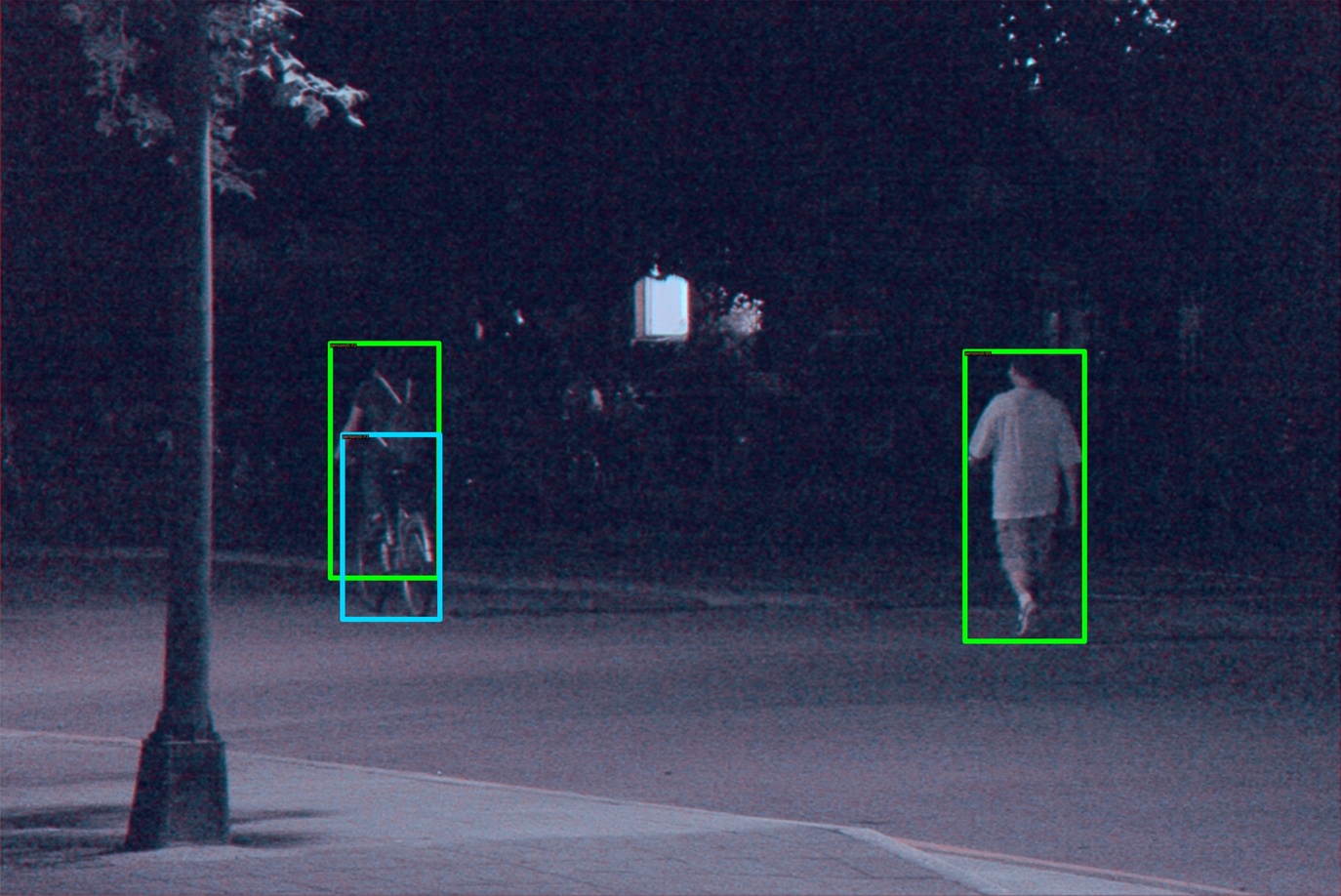}
\\
\smallskip
\includegraphics[width=38mm]{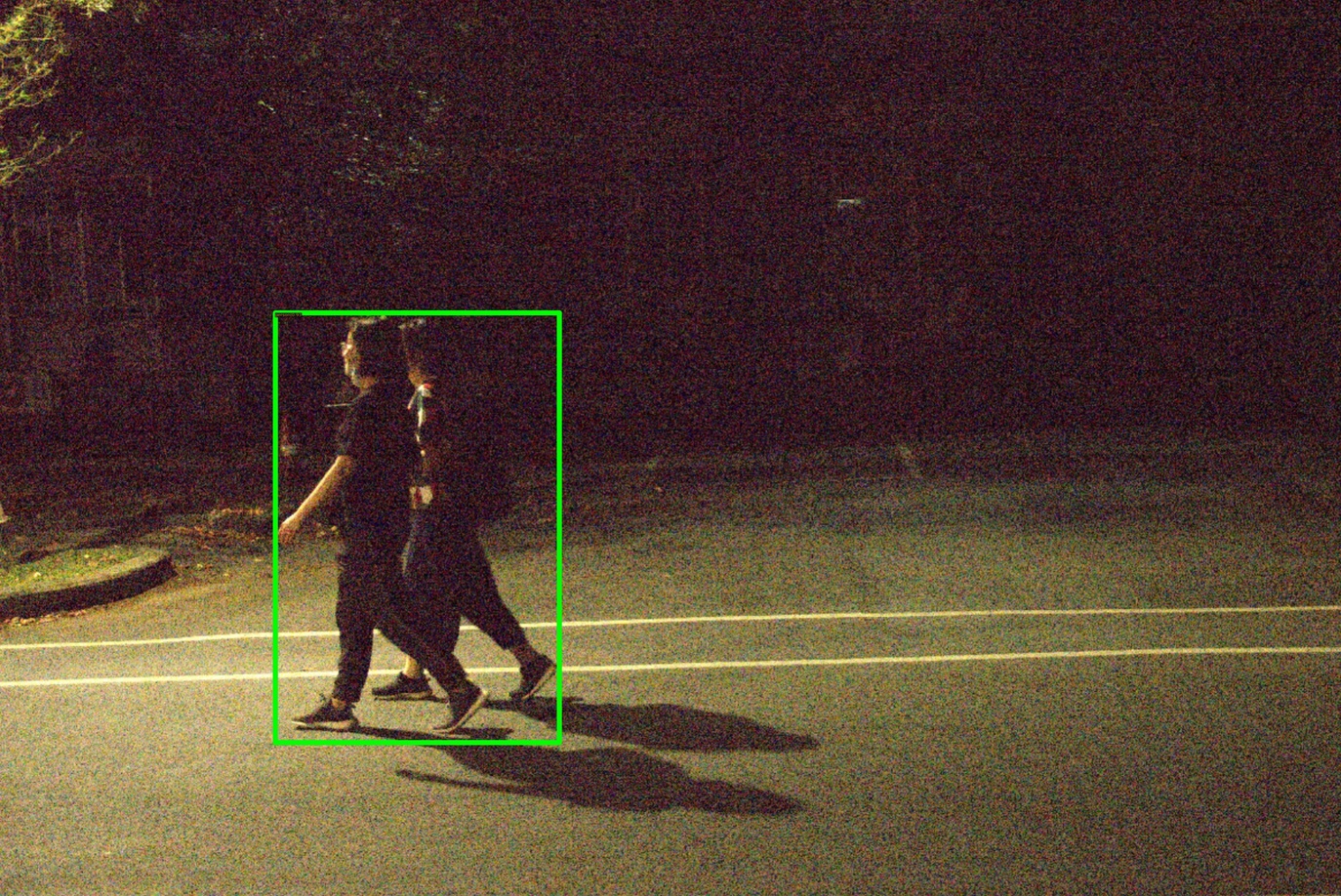}
\includegraphics[width=38mm]{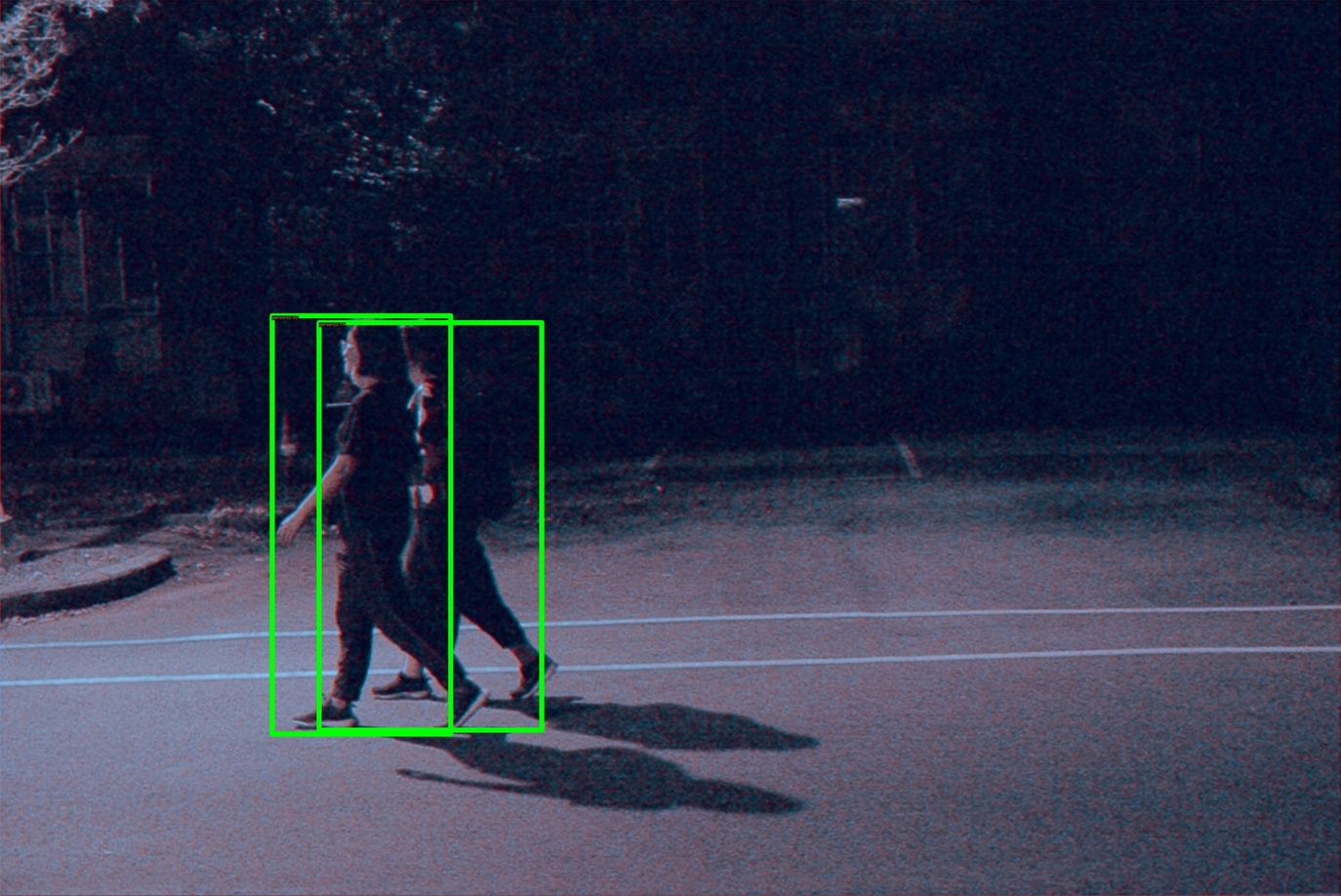}

\medskip

\includegraphics[width=80mm]{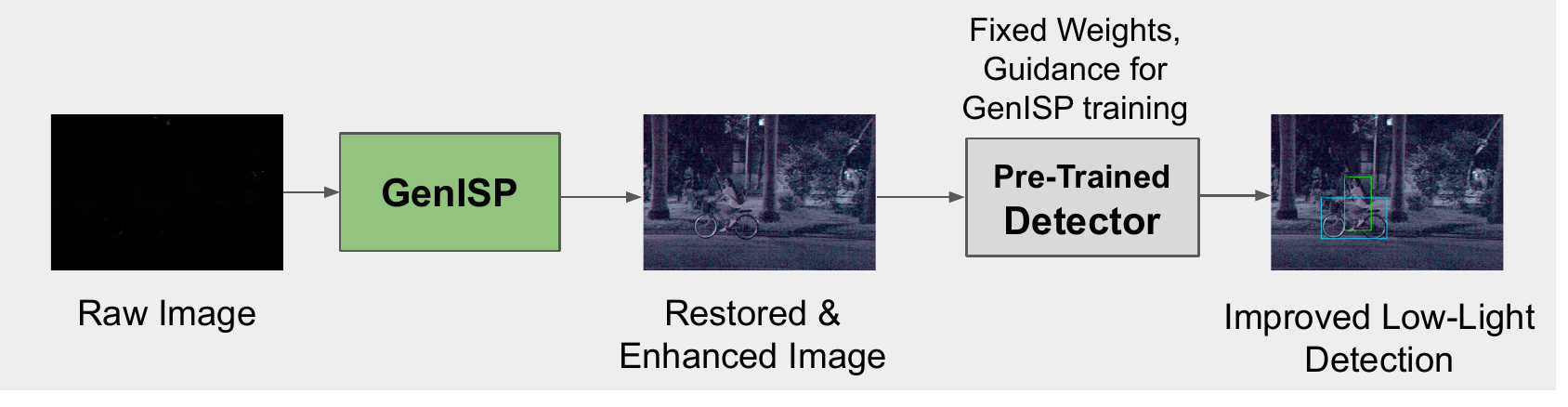}
\vspace{-1.5em}
\end{center}
\caption{The overview of our proposed method and visual results of low-light object detection with an off-the-shelf object detector trained on MS COCO \cite{COCO} using input processed by a traditional ISP pipeline (left) and our proposed neural ISP pipeline (right). }
\vspace{-2em}
\label{fig:first}
\end{figure}
Recently, low-light object detection received increased attention from researchers as a non-trivial problem in machine perception. Loh and Chan \cite{Exdark} have first shown that deep low- and normal-light features belong to different clusters, \textit{i.e.}, deep ConvNets do not learn to normalize data w.r.t. the lighting conditions. This shows that low- and normal-light data are fundamentally different and that low-light image perception is challenging, hence requires special considerations. Although there are modalities that overcome this limitation, visible-light cameras are cheaper, ubiquitous and perception using RGB images is relatively well-researched. Furthermore, enabling effective low-light perception using visible-light cameras can extend the functionality of already existing systems employing camera sensors.

Most studies in the computer vision community use image data processed by an in-camera traditional Image Signal Processor (ISP). However, Hong \etal \cite{Hong2021Crafting} has shown that, at least in low-light conditions, object detectors using raw sensor data perform better than detectors using sRGB data processed by a traditional ISP pipeline. As noted by Chen \etal \cite{SID}, traditional ISP pipelines are typically not tuned to process low-light data, and thus they fail in extreme low-light conditions. This observation might explain the result of Hong \etal \cite{Hong2021Crafting}. Although it is easy to leverage the advantages of raw data by training an object detector on a dedicated dataset, this strategy has limited applicability in real life. Most importantly, there is a lack of dedicated low-light object detection datasets using raw sensor data. Moreover, because raw image data is typically in a sensor-specific raw-RGB color space, a detector trained for a specific sensor is not likely to generalize to other camera sensors. 

An alternative approach is to process low-light data by a dedicated low-light restoration and enhancement method such as \cite{lamba2020towards,lamba2021restoring,SID}. However, most neural ISP methods use raw image data in the raw-RGB color space, and thus implicitly learn Color Space Transformation (CST) that maps the sensor-specific raw-RGB color space to a device-independent color space. Hence, they have limited generalization capability to camera sensors other than the one used during the training.
This is a major limitation in practice, as these models typically need paired well-lit ground-truth data that is difficult to collect. Moreover, most low-light restoration and enhancement methods produce images visually pleasant for the Human Visual System (HVS). As a consequence, they may not be optimal for machine vision.

In contrast with previous works, we propose to train a neural ISP pipeline, named GenISP, under the guidance of a pre-trained object detector. In this way, we make no assumptions about the processed image, and we do not need paired well-lit ground-truth data. Next, we propose to integrate CST matrices provided in raw data files into our proposed neural ISP pipeline to avoid implicit learning of a CST and for better generalization to unseen camera sensors. Moreover, we introduce expert knowledge into the neural ISP pipeline by proposing a two-stage color correction implemented by image-to-parameter network. We present extensive task-based experiments that show that our proposed method can outperform related restoration and enhancement methods while maintaining computational efficiency. Furthermore, we validate capability to generalize to unseen camera sensors, even when CST matrices are not available, and improving the performance of detectors other than the one employed in training.

\medskip
To summarize, our contributions are as follows:
\begin{itemize}
\itemsep0em 
    \item We propose a framework for training a neural ISP model for low-light image restoration and enhancement with an object detection dataset. Our minimal ISP pre-processing pipeline explicitly incorporates Color Space Transformation (CST) matrices available with raw files, instead of encoding CST implicitly. This helps improving the capability to generalize to unseen sensors and eliminate the need for re-training for each camera model.
    \item We propose a two-stage color correction implemented by two image-to-parameter modules: ConvWB and ConvCC. The two modules introduce expert knowledge about ISP and improve the detection results both when CST matrices are available and unavailable.  
    \item We validate in an extensive experimental study that once trained, the proposed model, GenISP, generalizes well to unseen datasets, camera sensors, brightness levels and object detectors. 
    \item We contribute a dataset of 7K raw images collected using two cameras, Sony RX100 (3.2K images) and Nikon D750 (4.0K), and bounding box annotations of \textit{people}, \textit{bicycles} and \textit{cars}. The dataset is publicly available for task-based benchmarking of future low-light image restoration and low-light object detection.
\end{itemize}

\section{Related Work}
Our work is at an intersection of low-light image restoration and enhancement, neural ISPs and low-light object detection. We first give an overview of related work in low-light object detection and motivation for using raw data. Next, we give an overview of deep ISPs that can be used for adapting raw data into a representation suitable for pre-trained off-the-shelf object detectors.

\subsection{Low-Light Detection}
\textbf{Low-light object detection from sRGB data.} 
In recent years, low-light face and object detection have received the attention of researchers as a separate problem \cite{Exdark,REG,poor_visibility_benchmark,morawski2021nod} rather than a subtask of object detection. Loh and Chan \cite{Exdark} have shown that ConvNets do not learn to normalize image data with respect to the lighting conditions, \textit{i.e.,} features extracted from low- and normal-light data belong to different data clusters. Morawski \etal \cite{morawski2021nod} have shown that low light conditions, based on the proportion of visible object edges, can be regarded as a spectrum of conditions ranging from extreme low-light conditions (difficult for machine perception) to non-extreme low-light conditions (relatively easy for machine perception). Further, Morawski \etal \cite{morawski2021nod} proposed to include an image enhancement module into the object detection framework to produce image representation optimal for machine perception. Similar difficulties exist in face detection. Liang \etal \cite{REG} proposed a detector based on generation of multi-exposure images from a single image. Wang \etal \cite{wang2021hla} proposed a joint high-low level adaptation to address high and low-level domain gaps by integrating low-level enhancement, darkening and feature adaptation. Moreover, the results of a comparative study presented by Wang \etal \cite{wang2021hla} suggests that enhancement-based methods improve detection results more than unsupervised domain adaptation methods. However, these prior works do not study the advantages of raw sensor data, a more robust image representation in low light conditions.
 
\textbf{Low-light object detection from raw data.} Recently, low-light object detection using raw sensor data has been proposed as a modality more robust against low illumination changes. As showed by Hong \etal \cite{Hong2021Crafting}, detectors using raw sensor data perform significantly better than detectors using sRGB data processed by a traditional ISP pipeline. However, raw sensor datasets are difficult to collect, store and annotate. To overcome this bottleneck, some works propose to leverage the abundance of labelled normal-light data \cite{Hong2021Crafting}\cite{sasagawa2020yolo}. Hong \etal \cite{Hong2021Crafting} proposed a framework synthesizing low-light raw sensor data by unprocessing normal-light sRGB data and a low-light recovery module. Sasagawa and Nagahara \cite{sasagawa2020yolo} proposed glue layers for domain adaptation by synthesizing darkened latent features from normal-light sRGB data processed by traditional ISPs. In contrast with these methods, we propose a neural ISP that adapts raw image data into representation optimal for machine cognition so that a pre-trained object detector can be used without any need for fine-tuning or re-training. 

\begin{figure}
\begin{center}

\begin{tabular}{c c c}
\includegraphics[width=20mm]{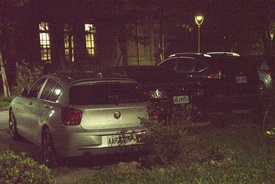}&
\includegraphics[width=20mm]{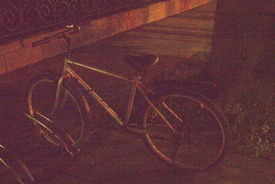}&
\includegraphics[width=20mm]{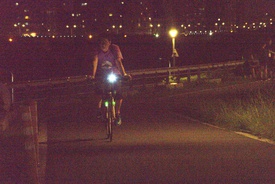}
\\
\includegraphics[width=20mm]{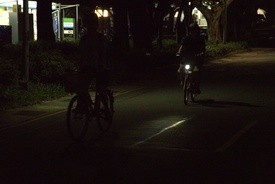}
&\includegraphics[width=20mm]{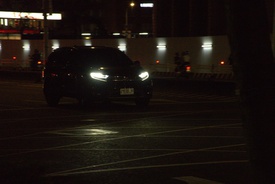}
&\includegraphics[width=20mm]{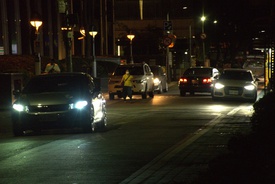}
\\
\includegraphics[width=20mm]{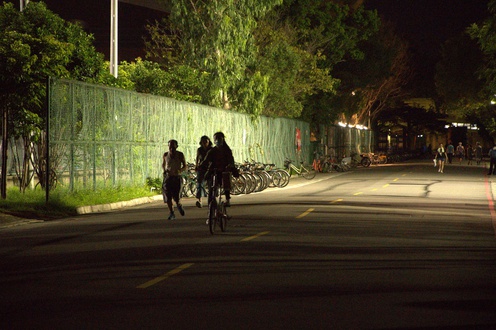}&
\includegraphics[width=20mm]{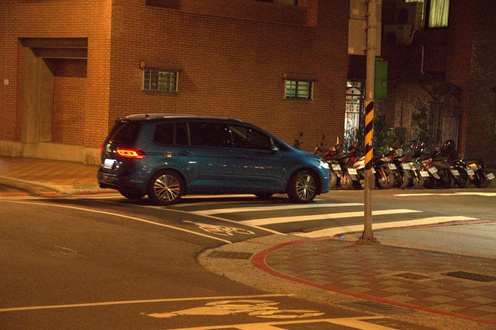}&
\includegraphics[width=20mm]{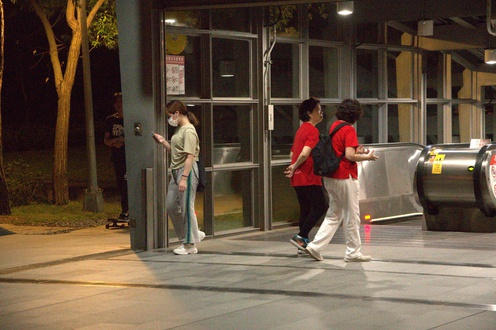}
\\
\end{tabular} 
\vspace{-1.5em}
\end{center}
\caption{Dataset variety. Or dataset consists of over 7K raw images collected using two cameras, Sony RX100 and Nikon D750, and bounding box annotations of \textit{people}, \textit{bicycles} and \textit{cars}. It shows a variety of low-light conditions in an unconstrained environment: ranging from extreme photon-limited conditions to less challenging conditions with artificial lighting.}
\label{fig:dataset_variety}
\vspace{-1.8em}
\end{figure}

\begin{figure*}[ht!]\vspace{-1.em}
\begin{center}
\includegraphics[width=13cm]{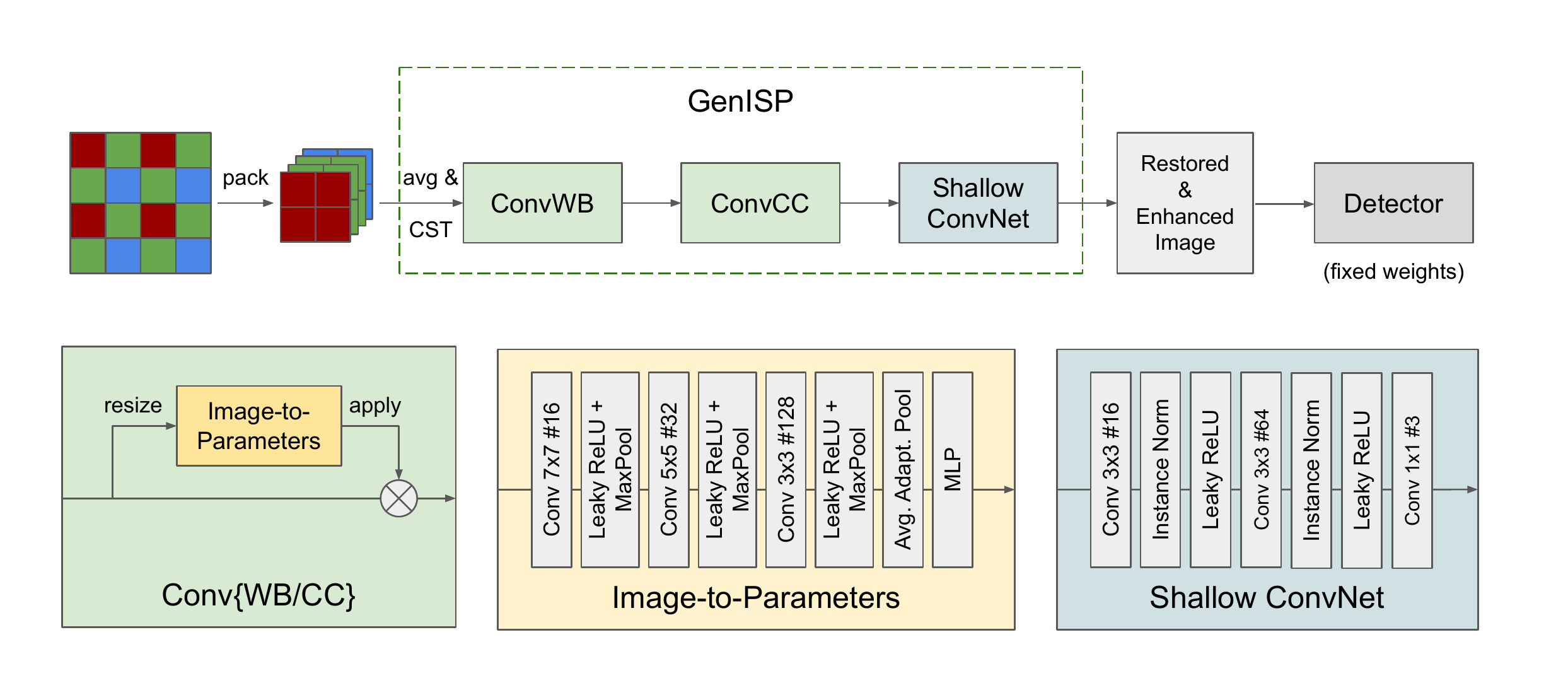}\vspace{-1.5em}
\end{center}
\caption{GenISP consists of a minimal ISP pipeline, two image-to-parameter modules and a shallow ConvNet. The proposed model, GenISP, is trained under the guidance of a pre-trained object detector with fixed weights without any paired or unpaired ground-truth well-lit data. The two proposed image-to-parameters modules ConvWB and ConvCC integrate expert knowledge inspired by traditional ISP pipeline and take a down-sized image as an input and regress parameters for global white balance (WB) and color correction (CC), respectively. Color-corrected image is then enhanced by a shallow ConvNet. The architecture makes for a small number of trainable parameters and a small count of operations that contribute to a short training time and efficient processing.}
\label{fig:method}
\vspace{-1.5em}
\end{figure*}

\subsection{Deep ISPs}
Deep learning methods have been applied to develop learning-based ISPs to process raw sensor data into sRGB images. A relatively simple deep learning model can replace traditionally hand-crafted ISP solutions composed of many separate processing steps, typically demosaicing, denoising, white balancing and color correction, contrast enhancement, to name a few. Moreover, each ISP pipeline step is a research topic on its own. Unlike traditional ISP pipelines that require expert knowledge and special efforts in tuning to the camera model, a learning-based solution can leverage paired training data to learn an optimal ISP that performs many processing steps in a joint way \cite{liu2020joint,xing2021end,qian2019rethinking,CameraNet}. Moreover, deep ISPs have been proposed for mobile cameras \cite{schwartz2018deepisp,ignatov2020replacing} to leverage the computational power of mobile phone platforms and close the gap between small mobile camera sensors and large DSLR camera sensors. Deep ISPs also have been proposed to learn joint image restoration and low-light image enhancement \cite{SID,lamba2020towards, lamba2021restoring}.

\textbf{Deep ISPs for low-light image restoration.} Traditional ISP pipelines are not well-suited for low-light and photon-limited conditions and often break down in those scenarios. Furthermore, traditional ISPs introduce errors and irreversible loss of information in images. Thus enhancement methods applied on images processed by traditional ISPs are not suitable for extreme low light conditions. Motivated by this, Chen \etal \cite{SID} proposed to use an end-to-end network to restore and enhance low-light images and contributed the See-in-the-Dark (SID) dataset of paired low-light and well-lit data of indoor and outdoor scenes. However, their proposed method requires a ground-truth amplification ratio applied to the input before processing data, which is not available in most real-life scenarios. Lamba \etal \cite{lamba2020towards} proposed LLPackNet for lightweight low-light image restoration with a neural amplifier to overcome the limitation of Chen \etal \cite{lamba2021restoring} and UnPack operation for up- and down-sampling data for improved color restoration. However, this neural amplification module is prone to introducing visually disturbing artifacts. Motivated by this, Lamba and Mitra \cite{lamba2021restoring} introduced an amplifier based on exponential binning that can be used with any restoration network. Lamba and Mitra \cite{lamba2021restoring} further proposed a light-weight architecture for image restoration by leveraging architectural parallelism. Lamba and Mitra \cite{lamba2021restoring} have also proposed improving Residual Dense Block that processes both rectified and non-rectified feature maps for a better information flow in extremely low-light images. 
However, most related methods propose to train the restoration model using data in the sensor-specific raw-RGB color space. In this setup, the model implicitly learns Color Space Transformation (CST) to map the sensor-specific color space to the device-independent sRGB color space. Thus, the generalizability of models trained in this setup is poor.  Furthermore, these works focus on improving perceptual quality and thus use datasets of paired low- and normal-light images. In contrast, we focus on improving task performance by generating images optimal for machine cognition, using low light raw data annotated with bounding boxes.

\section{Our Dataset}
We collected a dataset of outdoor images under low-light conditions using two cameras: Sony RX100 VII and Nikon D750. The dataset consist of 7K raw images (3.2K using Sony RX100 VII and 4.0K using Nikon D750) annotated with bounding boxes for 46K instances of \textit{people}, \textit{bicycles} and \textit{cars}. As shown in Fig. \ref{fig:dataset_variety}, the dataset presents a variety of low-light scenes in an unconstrained environment. Throughout the paper, we refer to the Sony subset as \textit{our Sony}, and to the Nikon subset as \textit{our Nikon}.
The dataset is publicly available at the project website for benchmarking of future methods targeting object detection in low-light conditions. 
\section{Proposed Method}
We propose a neural ISP called GenISP for low-light image restoration and enhancement, shown in Fig. \ref{fig:method}. Our neural ISP can be trained under the guidance of a pre-trained object detector to produce image representation optimal for machine cognition. It can adapt raw data in a sensor-specific color space to image representation optimal for the detector. By training under the guidance of an object detector with fixed weights, we eliminate any need for paired or unpaired low-light and normal-light ground-truth data, which can be difficult and laborious to collect.

Our motivation is to develop a neural ISP that can adapt raw sensor data into a representation suitable for any pre-trained off-the-shelf object detector. Thus, our GenISP is a minimal ISP that can help leveraging the advantages of raw data without fine-tuning the detector to a specific sensor.

The proposed method, shown in Fig. \ref{fig:method}, consists of a minimal pre-processing pipeline (packing and Color Space Transformation), a two-step color processing stage realized by image-to-parameter modules: ConvWB and ConvCC; and a non-linear local image enhancement by a shallow ConvNet.  Different from methods following SID \cite{SID}, our method does not require an externally specified amplification ratio or an external amplifier as in \cite{lamba2021restoring}. Rather, our method relies on the two modules ConvWB and ConvCC, as well as the design of the shallow ConvNet (by using InstanceNorm layers) to adjust brightness levels.  

\subsection{Color Space Transformation (CST)}
Most related restoration methods take packed raw data in the sensor-specific color space as an input and process data in an end-to-end fashion into an sRGB representation. In contrast, we propose to take advantage of Color Space Transformation (CST) matrices that map the sensor-specific color space (raw-RGB) to a device-independent color space (CIE XYZ). We first average green channels from the packed representation and then apply the available CST matrix. Hence, our model can better generalize to unseen camera sensors because it operates on data in a device-independent color space. CST matrices are available in most cameras and are included in the metadata of raw files. However, some camera manufacturers implement a minimal ISP pipeline, \textit{e.g.} in machine vision. The following modules, ConvWB and ConvCC, can help adapt the colour space in such a case.

Because typical image restoration methods use long-exposure images processed by a traditional ISP pipeline, they also implicitly use CST matrices by encoding CST in the network weights. In contrast with that, we propose to use CST matrices explicitly. In this way, our GenISP, trained on data from an arbitrary camera sensor, can be easily used with any other camera sensor.

\subsection{Neural Color Processing}
\label{sec:ncc}
We propose to integrate expert knowledge from traditional ISP pipeline by introducing a two-stage color processing consisting of white balancing and color correction. Inspired by Spatial Transformer Networks  \cite{NIPS2015_33ceb07b}, we propose two modules, ConvWB and ConvCC, that take a down-sized image as input and regress color transformation parameters, and apply the transformation to the image.

ConvWB predicts gain for each color channel of the input and controls to adjust global illumination levels and white balance of the image. Regressed weights $w_{ii}$ of a $3 \times 3$ diagonal WB matrix are applied to the image as in a traditional ISP pipeline:

\vspace{-0.4em}
\begin{equation}
\label{eq:wb}
    \begin{bmatrix}
    R' \\
    G' \\
    B' \\
\end{bmatrix} 
= 
    \begin{bmatrix}
    w_{11} & 0 & 0 \\
    0 & w_{22} & 0 \\
    0 & 0 & w_{33} \\
    \end{bmatrix}
    \begin{bmatrix}
    R \\
    G \\
    B \\
\end{bmatrix}.
\end{equation}
\vspace{-0.21em}

ConvCC takes the image corrected by ConvWB as an input and maps its color space to a color space optimal for a shallow ConvNet at the end of GenISP. 
Similarly to white balancing, color correction can be expressed as:

\vspace{-0.4em}
\begin{equation}
\label{eq:ccm}
    \begin{bmatrix}
    R' \\
    G' \\
    B' \\
\end{bmatrix} 
= 
    \begin{bmatrix}
    c_{11} & c_{12} & c_{13} \\
    c_{21} & c_{22} & c_{23} \\
    c_{31} & c_{32} & c_{33} \\
    \end{bmatrix}
    \begin{bmatrix}
    R \\
    G \\
    B \\
\end{bmatrix}.
\end{equation}
\vspace{-0.21em}

Although white balancing and color correction are realized by linear operators and can be combined by multiplying matrices in Eq. \ref{eq:wb} and \ref{eq:ccm}, we opt to keep the operations separate. We hypothesize that keeping the two modules separate can improve the optimization, as ConvWB matrix has less degrees of freedom and can perform a rough color correction. In this way, ConvCC module can be realized by a module with a smaller capacity.  

The architecture of the two modules is shown in Fig. \ref{fig:method} and is identical except for the number of units in the last layer of the MLP, 3 and 9, respectively.

\subsection{Training}
Unlike most low-light image enhancement and restoration methods, our neural ISP does not require paired or unpaired low-light and ground-truth normal-light data for training. Instead, our neural ISP is trained end-to-end under the guidance of an object detector. We pair a randomly initialized GenISP and a pre-trained object detector, and fix the detector's weights. In our experiments, we use a single-stage RetinaNet \cite{lin2017focal}. The total loss is composed of classification and regression loss:
\vspace{-0.29em}
\begin{equation}
\label{eq:loss}
    L_{total} = L_{cls} + L_{reg}.
\vspace{-0.29em}
\end{equation}
As for our experiments, we use RetinaNet as the detector for guiding GenISP during the training, so the classification loss is implemented by $\alpha$-balanced focal loss \cite{lin2017focal} and regression loss by smooth-L1 loss.
In contrast to prior works, we propose a minimal ISP for machine cognition rather than HVS. Thus, we make no constraints on the intermediate or processed image representation. In our experimental exploration, we found that intermediate loss terms to supervise the enhanced image $I_{enh}$ representation hurt the final performance of the ISP in a task-based detection evaluation. 

However, we observe than when trained with an object detection dataset without any constraints, GenISP reduces color saturation and loses colors not important for discrimination of objects in the dataset. If keeping the colors is important, an additional loss term can be employed:

\begin{equation}
\small
\label{eq:wb-loss}
\vspace{-0.25em}
    L_{wb} = \sum_{\forall (c_1, c_2) \in C} {|J_{enh}^{c_1} - J_{enh}^{c_2}|},C=\{(R, G), (R, B), (G, B)\},
\vspace{-0.25em}
\end{equation}
where $J_{enh}^{c}$ is an average intensity of the enhanced image $I_{enh}$ at color channel $c$. This loss term inspired by gray-world hypothesis encourages the model to balance the average intensity between each color channel. Similar losses were used in \cite{Zero-DCE} and \cite{sharma2021nighttime}. In this case, the total loss becomes:
\begin{equation}
\vspace{-0.25em}
\label{eq:loss-additional}
    L_{total} = L_{cls} + L_{reg} + \lambda_{wb} L_{wb}.
\vspace{-0.25em}
\end{equation}
We show the quantitative results of adding WB term in Fig. \ref{fig:color}.

The small number of trainable parameters and a small count of operations required to process an image (see Tab. \ref{tab:comparison_eff} in Sec. \ref{sec:results}) contribute to a short training time which makes fine-tuning for a custom sensor a feasible option.

\begin{table*}[ht]
\small
\begin{center}
\begin{tabular}{c | c | c c c c c}
\toprule 
 & Input & & & \multicolumn{3}{c}{Detection eval. ($\uparrow$)}  \\
 
Tested on & type & Method & Trained on & $AP_{50}$ & $AP_{75}$ & ${AP}$ \\

\hline

\multirow{10}{*}{Our Nikon} & \multirow{5}{*}{JPEG} & Baseline (Traditional ISP) & - & 41.6\% & 19.5\% & 21.2\%\\

 &  & Histogram Equalization & - & 36.9\% & 16.1\% & 18.5\%\\
 
 & & LIME \cite{LIME} & - & 41.2\% & 19.1\% & 20.7\%\\

 & & Zero-DCE \cite{Zero-DCE} & Our Sony & 40.7\% & 18.5\% & 20.5\%\\
 
 & & Zero-DCE \cite{Zero-DCE} & SICE \cite{cai2018learning} & 42.4\% & 19.7\% & 21.4\%\\

\cline{2-7}

 & \multirow{5}{*}{RAW}  & Lamba and Mitra \cite{lamba2021restoring} & SID Sony &  \multicolumn{3}{c}{did not generalize} \\
 
 & & Histogram Equalization & - &  40.3\% & 19.0\% & 20.5\%\\
 
  & & SID \cite{SID} & SID Sony &  44.5\% & 21.8\% & 22.9\%\\
  
 & & Lamba and Mitra \cite{lamba2020towards} & SID Sony &  45.5\% & 21.2\% & 23.0\%\\

 & & Our & Our Sony & \textbf{47.0\%} & \textbf{23.4\%} & \textbf{24.5\%} \\
\bottomrule
\end{tabular}
\vspace{-1.5em}
\end{center}
\caption{Cross-sensor (cross-dataset) extrinsic evaluation of enhancement and restoration methods. We use Average Precision ($AP$) to evaluate the detection performance at different $IoU$ thresholds, \textit{i.e.}, $IoU=.50$, $.75$ and $0.50:0.95$ denoted by $AP_{50}$, $AP_{75}$ and ${AP}$, respectively. We hypothesize that the poor performance of methods using JPEG data is caused by propagation of errors made by the traditional ISP pipeline. Because there are currently no datasets with ground-truth well-lit images and bounding box annotation at the same time, we compare all methods in a cross-dataset setup for a fair comparison. Baseline denotes traditional ISP pipeline using RawPy \cite{rawpy}.}
\label{tab:quant_comparison}
\vspace{-0.7em}
\end{table*}

\begin{table*}[ht]
\small

\begin{center}
\begin{tabular}{c | c | c c c c c}
\toprule 
 & Input & & & & & \\
 
Tested on & type & Method & Trained on & Parameters (M) ($\downarrow$) & GMACs ($\downarrow$)  & AP ($\uparrow$) \\

\hline

\multirow{4}{*}{Our Nikon} &  JPEG & Baseline (Traditional ISP) & - & - & - & 22.2\%\\

\cline{2-7} 

&  \multirow{3}{*}{RAW} & SID \cite{SID} & SID Sony &  7.7 & 562 & 22.9\%\\

 & & Lamba and Mitra \cite{lamba2021restoring} & SID Sony &  0.78 & \textbf{60} & 23.0\%\\

 & & Our & Our Sony & \textbf{0.12} & \underline{ 61} & \textbf{24.5\%} \\
\bottomrule
\end{tabular}
\vspace{-1.5em}
\end{center}
\caption{Comparison to the state-of-the-art neural ISP methods for low-light image restoration. Detection performance is reported in a cross-sensor (cross-dataset) setup. Traditional ISP pipeline is implemented using RawPy \cite{rawpy}.}
\label{tab:comparison_eff}\vspace{-1.5em}
\end{table*}
\section{Experimental Results}
\label{sec:results}

Since we aim to improve detection results in challenging low-light conditions, we evaluate our neural ISP in an extrinsic task-based evaluation using standard object detection metrics (mean Average Precision, mAP). In our paper, we focus on developing a neural ISP that produces images optimal for machine cognition rather than the Human Visual System (HVS). In contrast with most related work, we do not impose any constraints on how the adapted image should look. Thus, we do not assess perceptual quality or similarity to normal-light reference data, even if available.

We train using Sony RX100 (Bayer sensor, $5496 \times 3672$) from our dataset and report our results on the same sensor as well as other camera sensors for testing cross-sensor generalization. The other sensors are: Nikon D750 (Bayer sensor, $3968 \times 2640$) in our dataset, Sony \textalpha7S II (Bayer sensor, $4256 \times 2848$) in the SID dataset \cite{SID} and Nikon D3200 DSLR camera (Bayer sensor, $6034 \times 4012$) in the PASCALRAW \cite{omid2014pascalraw} dataset. 

Bounding box ground truth is available for our Sony, our Nikon and PASCALRAW \cite{omid2014pascalraw}. Different to these datasets, the See-in-the-Dark (SID) dataset \cite{SID} is an image restoration benchmark and has no image annotations. Instead, it consists of short- and long-exposure paired raw image data, where long-exposure data is used as ground-truth for image restoration. We follow Lamba and Mitra \cite{lamba2021restoring} and for each short-exposure image, we detect objects from its long-exposure ground-truth image using Faster R-CNN \cite{Ren_2017}, and retain predictions with a confidence score over 70\% for pseudo ground truth. A similar evaluation methodology was also used by Sasagawa and Nagahara \cite{sasagawa2020yolo}.

\subsection{Implementation Details}
We implement our method using Open MMLab Detection Toolbox \cite{mmdetection} and PyTorch \cite{paszke2019pytorch} running on a single Tesla-V100 32GB GPU. We train the proposed restoration model under the guidance of RetinaNet \cite{lin2017focal} with ResNet-50 \cite{he2016deep} as the backbone. We freeze all parameters of RetinaNet and optimize GenISP using Adam \cite{adam}. We train for 15 epochs, and set the batch size to 8, learning rate to $\expnumber{1}{-2}$ initially and decrease it to $\expnumber{1}{-3}$ and $\expnumber{1}{-4}$ at the 5th and 10th epoch, respectively. During training and testing, we resize the images to a maximum size of $1333 \times 800$ and keep the image aspect ratio. In ConvWB and ConvCC, we resize input to $256 \times 256$ using bilinear interpolation. In the case of raw data, we first pack raw data into 4 color channels (RGGB) before resizing. Besides geometric transformations for object detection, we apply brightness and contrast augmentation to the training data. Training using our Sony RX100 dataset in this setup takes only 1.5 hours. The short training time as well as low number of epochs is associated with a low number of trainable parameters (0.12M) and a simple architecture. All experiments in the paper use a model trained on Our Sony in the setup detailed above.

\subsection{Quantitative Comparison}
First, we evaluate and compare the proposed method and related low-light enhancement and restoration methods. We perform an extrinsic, task-based evaluation to study how the restoration and enhancement modules improve the performance of an off-the-shelf pre-trained object detector. A similar evaluation methodology was used in \cite{wang2021hla}. 

For a fair comparison, we perform the study in a cross-sensor setup and fine-tune using our Sony if possible. Because our framework uses object detection annotation rather than paired short- and long-exposure data, it was not possible to compare the models directly using the same training data. For methods using sRGB data, denoted by JPEG, we process raw data using an open-source ISP pipeline RawPy \cite{rawpy}, showed in Fig. \ref{fig:dataset_variety}. We used RawPy rather than in-camera ISPs, as in-camera ISPs of our Sony and Nikon implement geometric transformation, and because of that raw and JPEG images using in-camera ISP would differ. As for SID \cite{SID}, it requires an externally set pre-amplification ratio, \textit{e.g.,} based on the proportion of exposure times of short-exposure low-light image and long-exposure ground-truth image, not available in our dataset. We experimentally determined the pre-amplification ratio to be 100 and apply it to all images in our dataset. For histogram equalization, we separately equalized each channel of images processed by a traditional ISP pipeline (denoted by JPEG) and raw images (denoted by RAW). As for raw histogram-equalized images, we averaged green channels so that the resulting image can be processed by an unaltered object detector.

In Tab. \ref{tab:quant_comparison} we show the results of the evaluation using a pre-trained RetinaNet \cite{lin2017focal} with ResNet-50 \cite{he2016deep} as the backbone. The performance of the detector using JPEG data deteriorates when using low-light enhancement methods. We hypothesize that this is because enhancement methods further propagate errors made by the traditional ISP pipeline, which is not well-suited to processing low-light images. On the other hand, learning-based ISP methods (denoted by RAW) improve the results of the detector.

Further, in Tab. \ref{tab:comparison_eff} we show comparison to other state-of-the-art neural ISP methods for low-light imaging in terms of performance and computational efficiency. GenISP improves the detection results while maintaining attractive computational efficiency, comparable to Lamba and Mitra \cite{lamba2021restoring} that targeted light-weight real-time restoration.

\subsection{Cross-Dataset and -Sensor Generalization}

Next, we validate the generalization capabilities of our proposed method in a cross-dataset and cross-sensor setup.  In Tab. \ref{tab:cross_dataset} we report the results of the task-based evaluation: we trained on our Sony RX100 dataset and tested on datasets as reported in the table. Our proposed method improves the detection results across all tested datasets (sensors) in both low-light and normal-light conditions. Thus, our proposed method, once trained on a low-light dataset, can be used with different sensors and is not constrained to low-light conditions. 
\begin{table}[h]
\small

\vspace{-0.5em}
\begin{center}
\begin{tabular}{ c | c c c}
\toprule
 & & \multicolumn{2}{c}{AP ($\uparrow$)} \\
 & & Baseline & Ours \\
\hline
 \multirow{3}{*}{Low-light} & Our Sony & 21.1\% & \textbf{25.7\% }\\
\cline{2-4}
 & Our Nikon & 22.2\% & \textbf{24.5\%} \\
 & SID Sony \cite{SID} & 21.2\% & \textbf{22.6\%} \\
\hline
 \multirow{1}{*}{Normal-light}  & PASCALRAW \cite{omid2014pascalraw} & 37.4\% & \textbf{39.6\%} \\
\bottomrule
\end{tabular}
\vspace{-1.5em}
\end{center}
\caption{Generalizability of our proposed method. We pair our proposed model trained on our Sony RX100 dataset with a pre-trained off-the-shelf object detector and report object detection results. Our proposed method improves detection results across all tested sensors in both low- and normal-light conditions.
Baseline denotes traditional ISP pipeline using RawPy \cite{rawpy}. }
\label{tab:cross_dataset}
\vspace{-1.5em}
\end{table}

\begin{figure*}[ht!]
\begin{center}


\includegraphics[width=40mm]{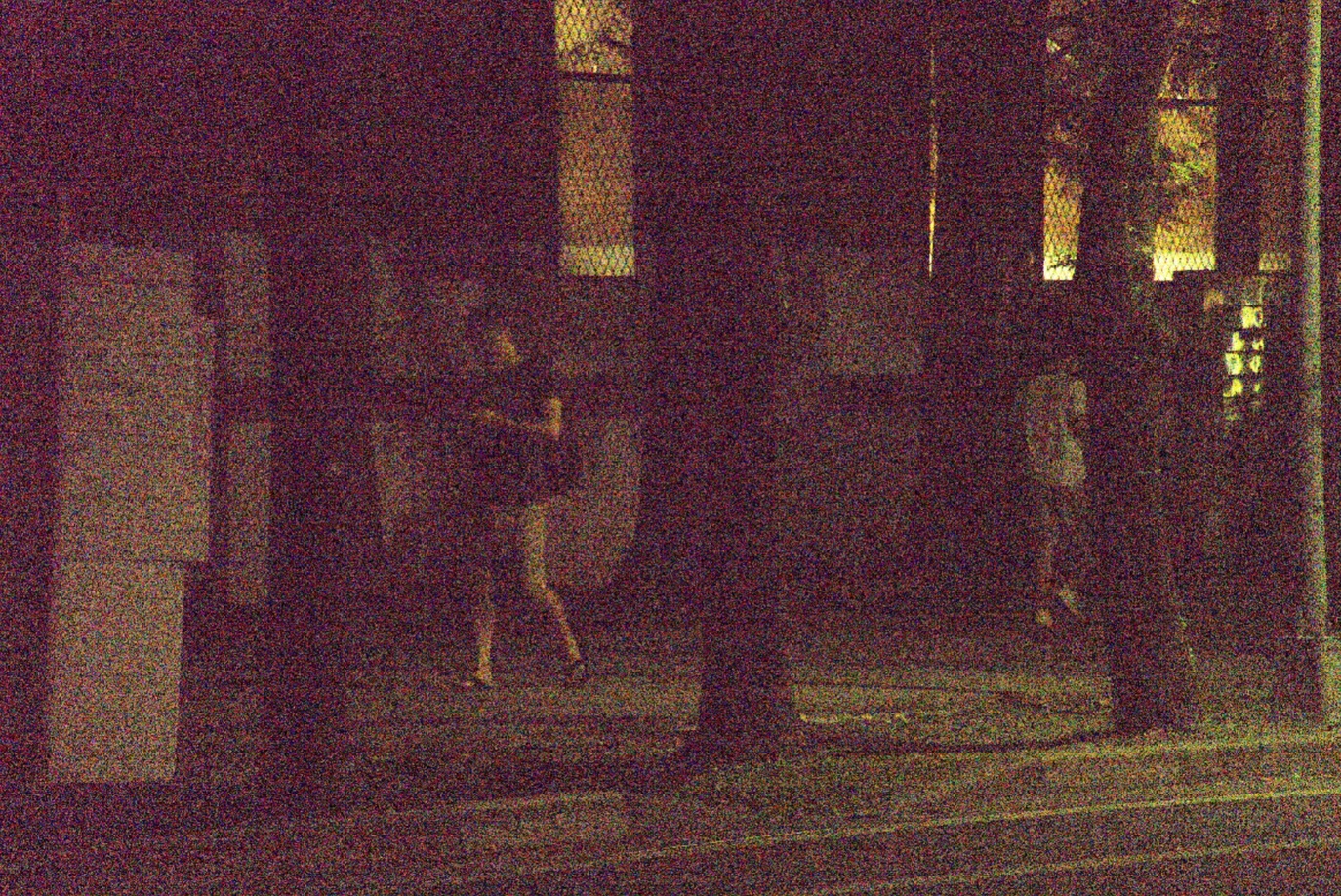} \
\includegraphics[width=40mm]{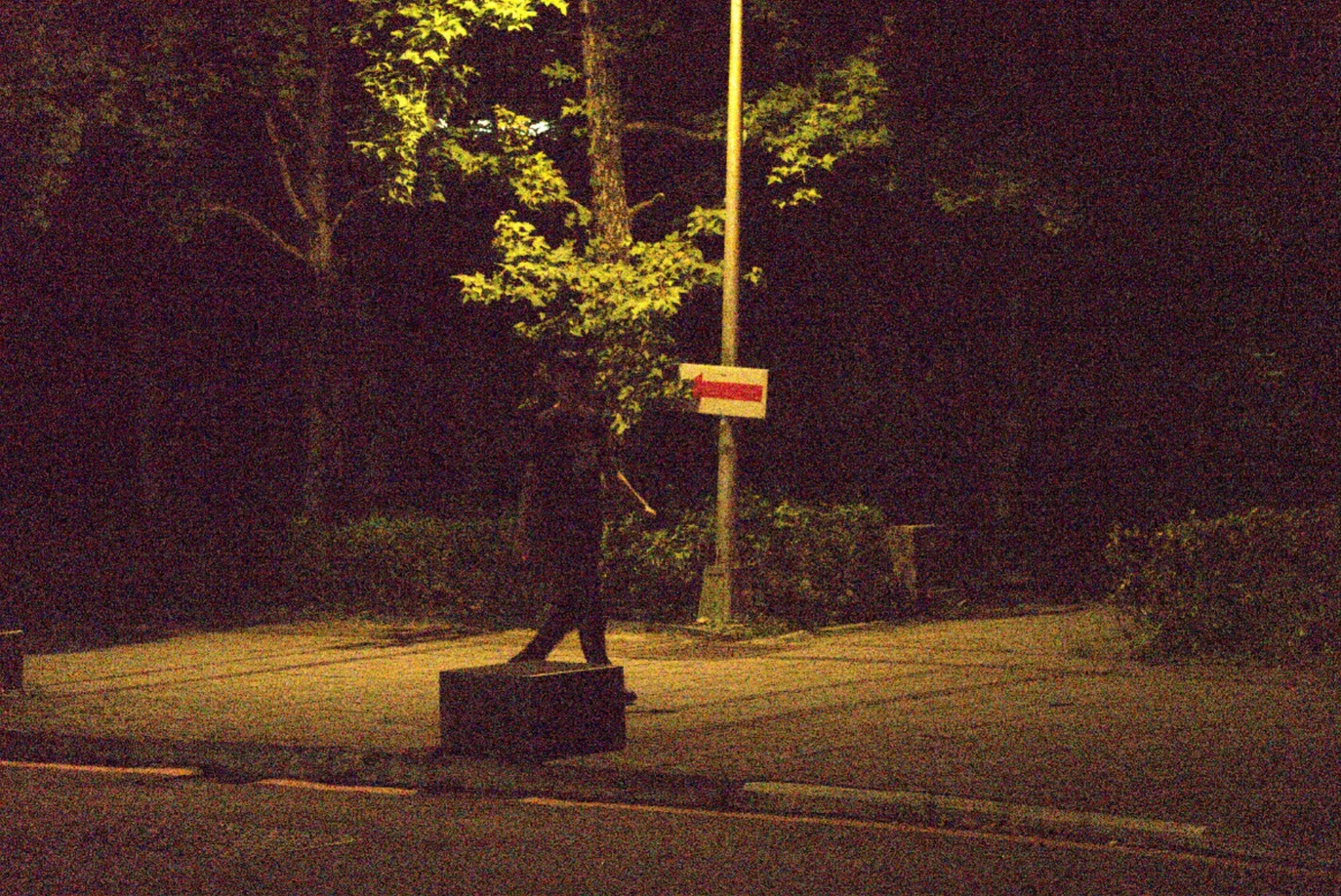} \ 
\includegraphics[width=40mm]{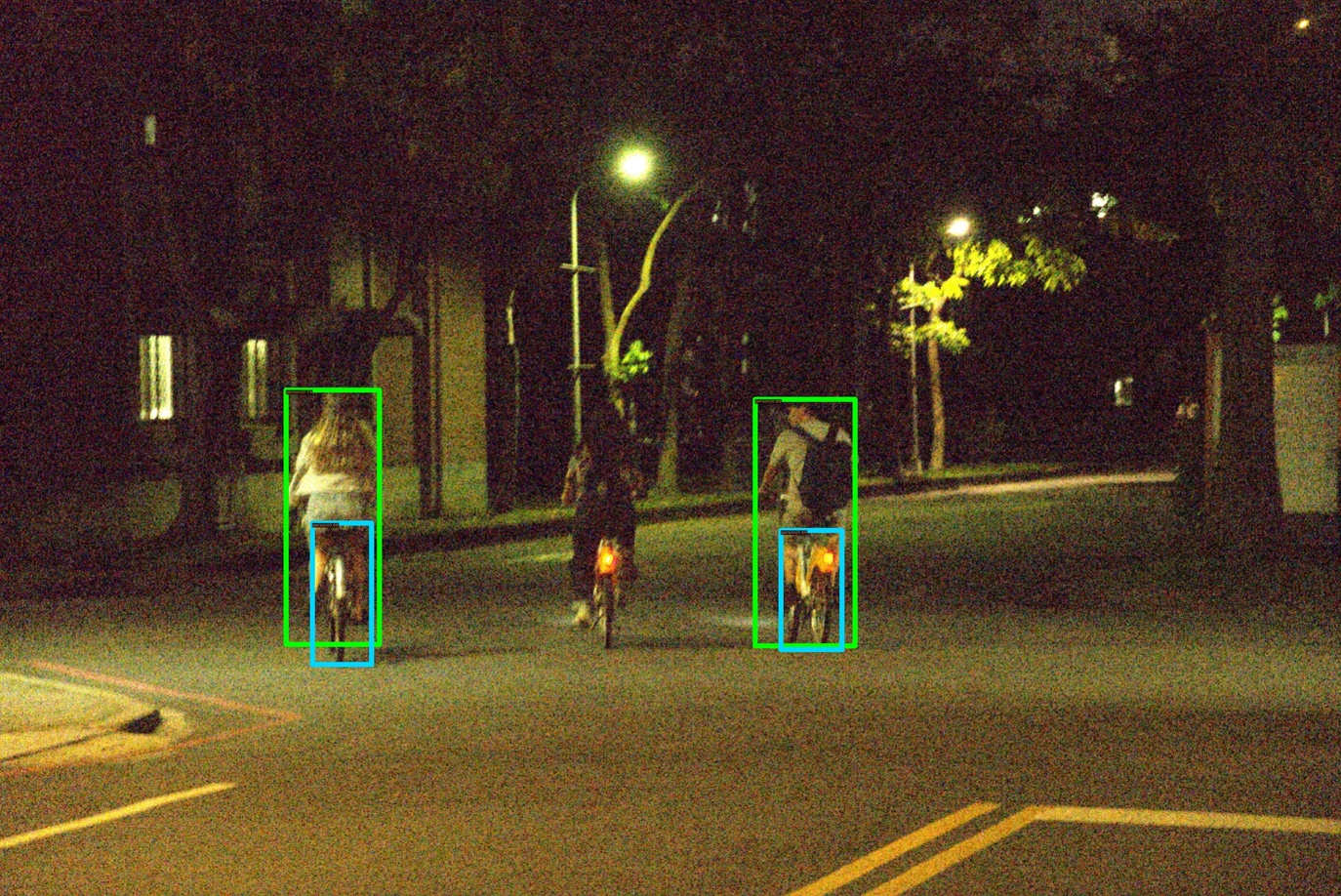} \ 
\includegraphics[width=40mm]{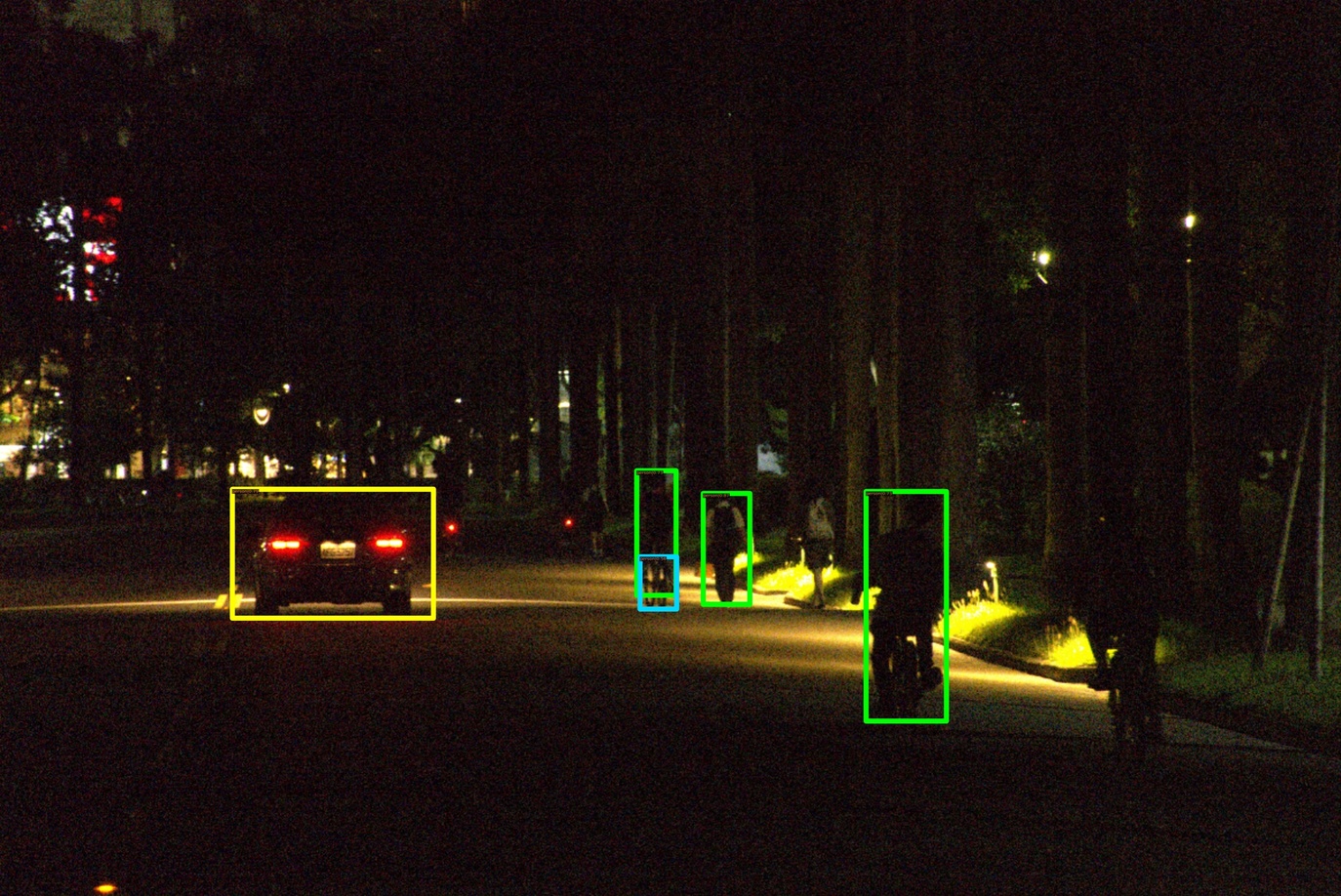}  
\smallskip
\includegraphics[width=40mm]{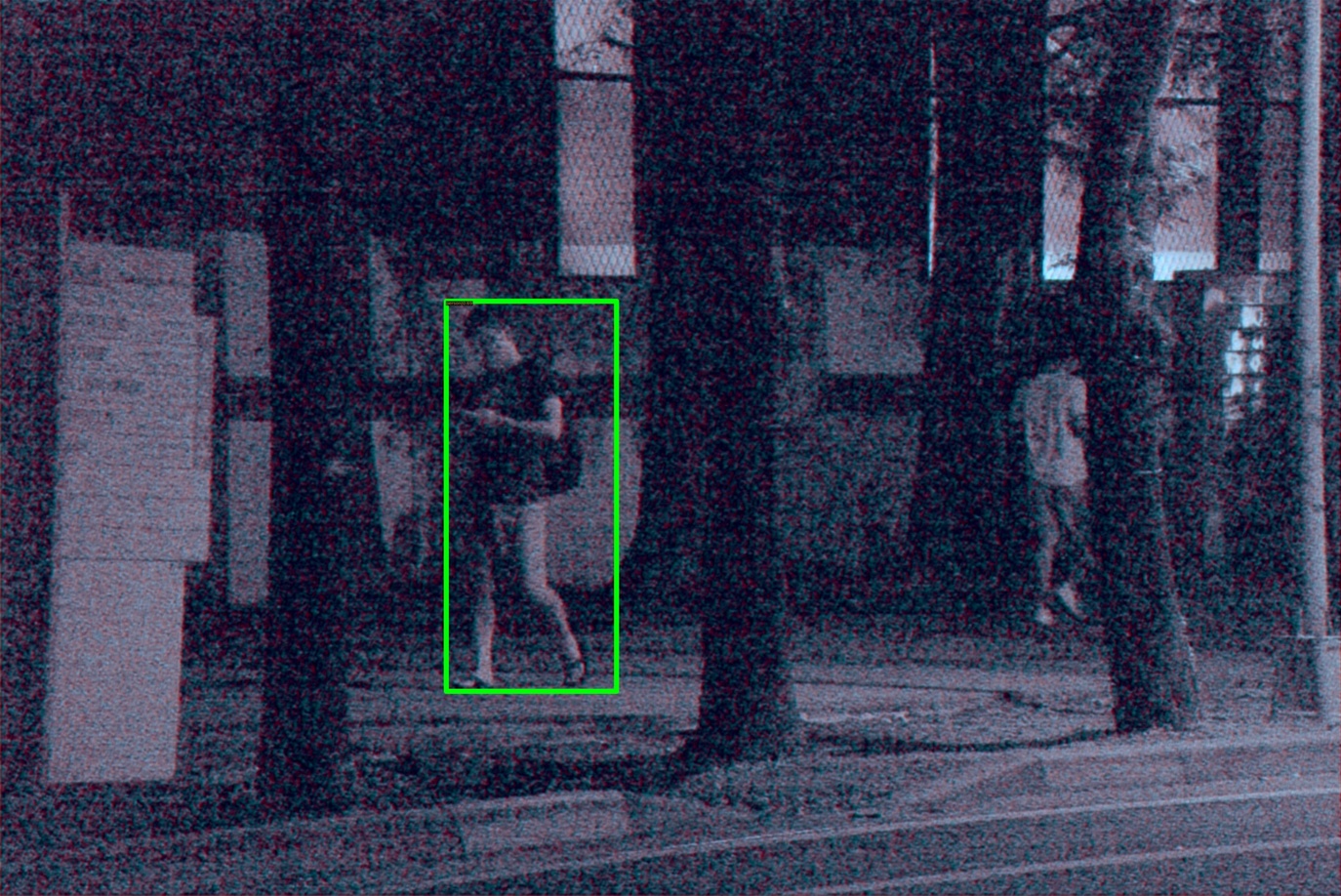} \
\includegraphics[width=40mm]{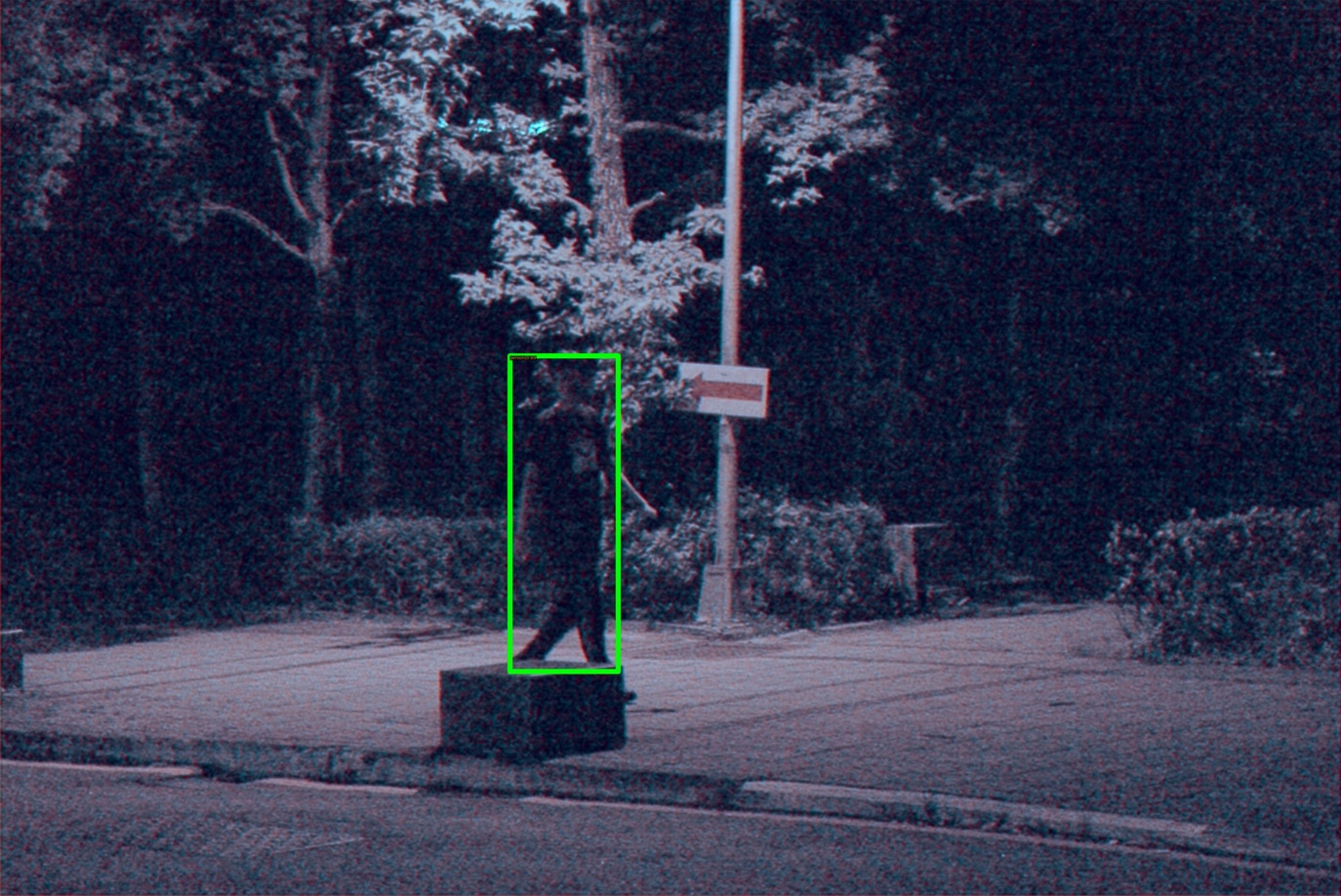} \
\includegraphics[width=40mm]{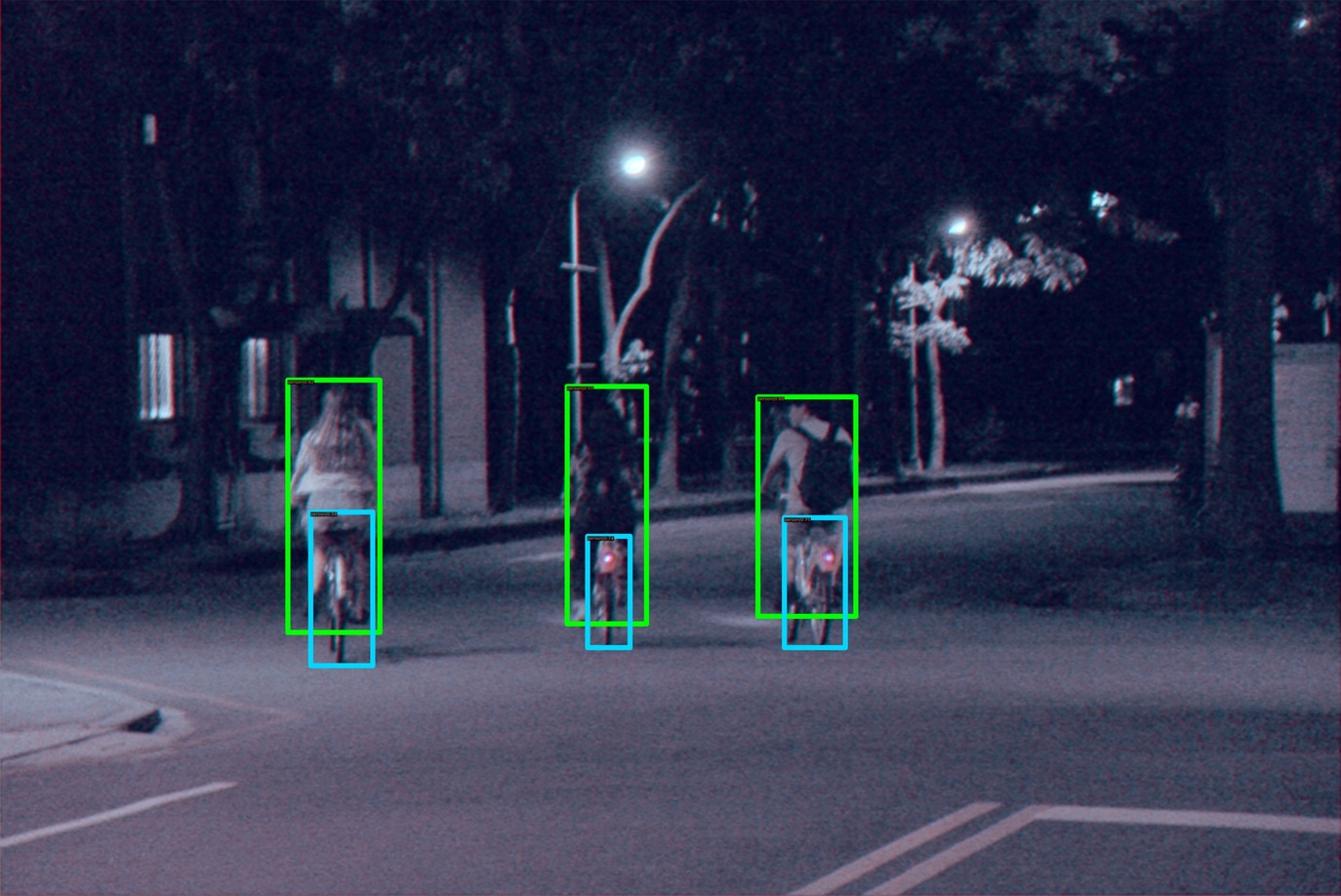} \
\includegraphics[width=40mm]{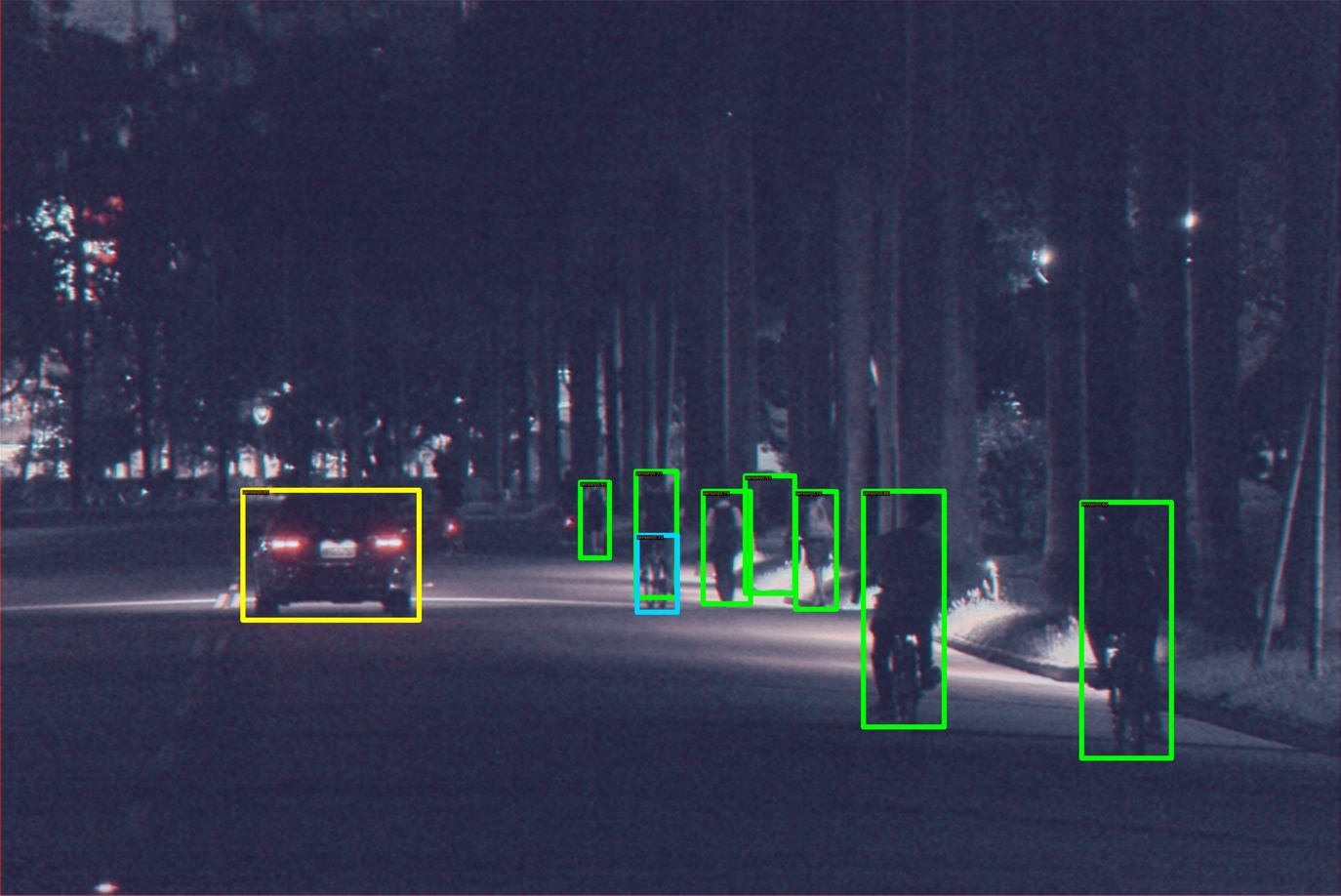}
\\

\vspace{-1.5em}
\end{center}
\caption{Qualitative results. Paired with a pre-trained object detector, our proposed neural ISP improves the detection results in challenging low-light conditions. Top row: traditional ISP + RetinaNet (pre-trained on MS COCO \cite{COCO}), bottom row: our proposed neural ISP + the same RetinaNet checkpoint (MS COCO \cite{COCO}).}
\label{fig:bb_imp}
\vspace{-1.5em}
\end{figure*}


\subsection{Cross-Detector Generalization}

To validate that our proposed method, once trained under the guidance of an object detector, can improve the detection results of many object detectors, we perform a cross-detector evaluation. We trained our proposed restoration method under the guidance of RetinaNet \cite{lin2017focal} with ResNet-50 \cite{he2016deep} and tested it when paired with other detectors. As can be seen in Tab. \ref{tab:cross_model}, the detection performance is generally improved across all tested detectors.

\begin{table}[ht]
\small
\begin{center}
\begin{tabular}{ c | c c c}
\toprule
 & & \multicolumn{2}{c}{AP ($\uparrow$)} \\
 & & Baseline & Ours \\
\hline
 \multirow{1}{*}{Two-stage} & Faster R-CNN \cite{Ren_2017} & 20.6\% & \textbf{25.1\%} \\
\hline
 \multirow{3}{*}{Single-stage} & FCOS \cite{tian2019fcos} & 18.7\% & \textbf{19.2\%} \\
 & RetinaNet \cite{lin2017focal} & 21.1\% & \textbf{25.7\% }\\
 & PAA \cite{paa-eccv2020} & 21.8\% & \textbf{25.8\%} \\
\bottomrule
\end{tabular}
\vspace{-1.5em}
\end{center}
\caption{Cross-model evaluation of our proposed method. Once trained under the guidance of an object detector (here RetinaNet \cite{lin2017focal}), our proposed module can be paired with a variety of detectors to improve detection results under low-light conditions. Baseline denotes traditional ISP pipeline using RawPy \cite{rawpy}.}
\vspace{-1.5em}
\label{tab:cross_model}
\end{table}

\begin{table}[ht]
\footnotesize
\begin{center}
\begin{tabular}{ c c c c c}
\toprule
 & \scriptsize Backbone & \scriptsize \# Param. (M) ($\downarrow$) & \scriptsize GFLOPs ($\downarrow$) & AP ($\uparrow$) \\
\hline
Baseline & \scriptsize ResNet-50 \cite{he2016deep} & 37.96 & 61.22 & 21.1\% \\
Ours & \scriptsize ResNet-50 \cite{he2016deep} & +0.12 & +3.3 & \textbf{25.7\%} \\
\hline
Baseline & \scriptsize ResNet-101 \cite{he2016deep} & 56.95 & 80.7 & 22.5\% \\
Ours & \scriptsize ResNet-101 \cite{he2016deep} & +0.12 & +3.3 & \textbf{25.9\%} \\
\bottomrule
\end{tabular}
\vspace{-1.5em}
\end{center}
\caption{Effect of using a deeper backbone in the detector on detection results under low-light conditions. Our proposed method is a more effective strategy for improving detection results than using a deeper backbone. Baseline denotes traditional ISP pipeline using RawPy \cite{rawpy}.}
\label{tab:cross_depth}
\vspace{-1.3em}
\end{table}

Further, in Tab. \ref{tab:cross_depth}, we study the effect of using a deeper backbone in the detector. A deeper backbone can extract higher-quality features and provide more robustness due to the higher model capacity. In contrast with this strategy, we propose to use raw image data and restore it using a neural ISP, so that the backbone can extract features from a higher-quality input. As witnessed by higher mAP values, our proposed restoration module improves the detection results more than using a deeper backbone. Additionally, it is a more effective strategy, as it adds less complexity and fewer additional model parameters. 

\subsection{Ablation Study}
 Finally, to validate the need of all proposed modules in the proposed method, we perform an ablation study in a cross-sensor setup. We train 4 variations of the proposed restoration model on our Sony and test on the SID Sony \cite{SID}. We report the results of this study in Tab. \ref{tab:ablation}. 
 
\begin{table}[h]
\footnotesize

\vspace{-0.5em}
\begin{center}
\begin{tabular}{c c c | c c}
\toprule
Shallow & & & \multicolumn{2}{c}{AP ($\uparrow$)} \\
ConvNet & ConvWB & ConvCC & w/ CST mtx. & w/o CST mtx. \\
\hline
  &   &   & 21.2\% & - \\
\checkmark &   &   & 21.7\% & 17.5\% \\
\checkmark & \checkmark &   & 21.2\% & 19.2\% \\
\checkmark &   & \checkmark & 21.2\% & 20.7\% \\
\checkmark & \checkmark & \checkmark & \textbf{22.6\%} & \textbf{21.2\%} \\
\bottomrule
\end{tabular}
\vspace{-1.5em}
\end{center}
\caption{Ablation study in two setups: with and without Color Space Transform (CST) matrix. CST matrices are provided in metadata of raw files. However, camera manufacturers sometimes implement a minimal ISP pipeline without CST. In such case, our method is also effective at improving detection results. }
\label{tab:ablation}
\vspace{-1.5em}
\end{table}

 When Color Space Transform (CST) matrices are available, our proposed modules ConvWB and ConvCC do not improve the detection results unless employed together. We hypothesize that, in this setup, color correction is less obvious and while ConvWB is not enough for improving the detection results, ConvCC has too many degrees of freedom to accurately predict color correction parameters. When employed together, ConvWB and ConvCC improve the detection results. On the other hand, when CST matrices are not available, ConvWB and ConvCC improve the detection results when employed together or separately. Overall, our proposed modules can improve the detection results and are more important when CST matrices are unavailable, which is the case when the camera manufacturer implements only a minimal ISP, \textit{e.g.}, in machine vision.

\subsection{Qualitative Results}
\begin{figure}[t!]
\begin{center}

\includegraphics[width=38mm]{fig/bb_imp/cstnet/DSC01544.JPG}
\includegraphics[width=38mm]{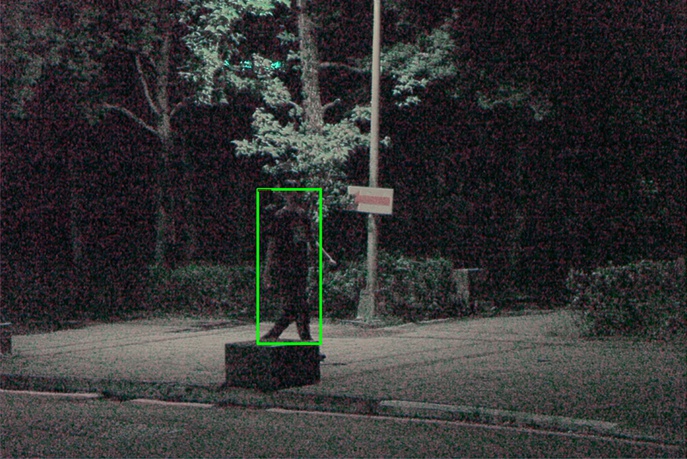}
\smallskip
\includegraphics[width=38mm]{fig/bb_imp/cstnet/DSC02156.JPG}
\includegraphics[width=38mm]{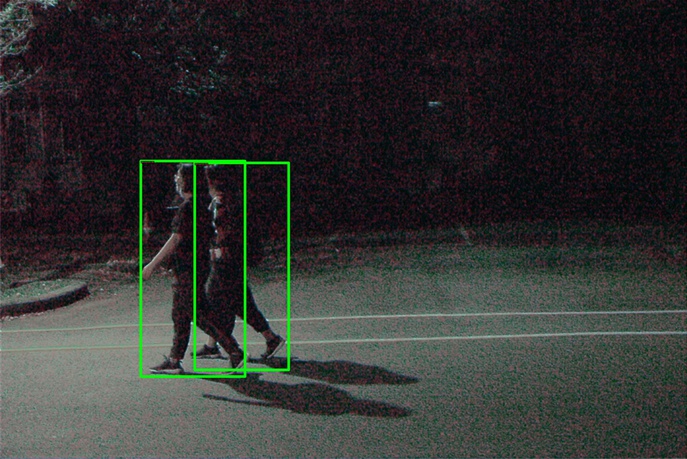}

\vspace{-1.5em}
\end{center}
\caption{ Results of GenISP without (left) and with (right) gray-world hypothesis loss in Eq. \ref{eq:wb-loss}.  }
\vspace{-2em}
\label{fig:color}
\end{figure}
In Fig. \ref{fig:bb_imp}, we show qualitative results of our method by visualizing detection results with confidence score over $0.5$. We pair our GenISP with an object detector, RetinaNet \cite{lin2017focal} with ResNet-50 \cite{he2016deep} as the backbone trained on MS COCO \cite{COCO}. Our GenISP improves the detection performance by improving confidence scores and object localization quality. 

In comparison with the traditional ISP, GenISP reduces the saturation of the processed images. Strong colors important for discrimination in our dataset, such as red car and bike lights, are still clearly present in the produced images. We hypothesize that color information is not crucial for discrimination in our dataset which is limited to a small number of classes. To confirm the hypothesis, we test the baseline model (pre-trained RetinaNet) on our Sony (JPEG) and its grayscale version, and obtained the following results: 21.1\% vs 21.8\%, respectively. A possible explanation for this improvement is that averaging color channels reduces noise. However, if retaining color information is important for an application, GenISP can be trained with an additional gray-world hypothesis loss term as in Eq. \ref{eq:wb-loss}, as shown in Fig. \ref{fig:color}.

\section{Conclusion}\label{sec:conclusion} 
We presented a new neural ISP method GenISP for low-light image restoration. Our proposed method processes raw sensor image data into sRGB image representation optimal for machine cognition rather than the Human Visual System (HVS). GenISP is trained under the guidance of a pre-trained object detector to avoid making any assumptions about how the enhanced image should look and eliminate the need for paired low- and normal-light ground-truth data. Moreover, in contrast to related works, we propose to take advantage of Color Space Transform (CST) matrices available in raw files and perform restoration and enhancement in a device-independent color space, instead of implicitly encoding CST in the network weights and degrading the generalization capacity of the trained model to unseen sensors. Additionally, we introduced expert knowledge about color correction into the proposed method by employing image-to-parameters modules that can also improve the performance in case CST matrices are unavailable. We performed extensive experiments to validate that the proposed method can improve detection results in cross-sensor and cross-detector setups. Finally, we contributed a raw image object detection dataset publicly available for task-based benchmarking of future low-light image restoration and low-light object detection.
\\
\textbf{Acknowledgements}: This work was supported in part by the Ministry of Science and Technology, Taiwan, under Grant MOST 110-2634-F-002-051 and Qualcomm Technologies, Inc. We are grateful to the National Center for High-performance Computing.


{\small
\bibliographystyle{ieee_fullname}
\bibliography{047}

\begin{thebibliography}{10}\itemsep=-1pt

\bibitem{cai2018learning}
Jianrui Cai, Shuhang Gu, and Lei Zhang.
\newblock Learning a deep single image contrast enhancer from multi-exposure
  images.
\newblock {\em IEEE Transactions on Image Processing}, 27(4):2049--2062, 2018.

\bibitem{SID}
Chen Chen, Qifeng Chen, Jia Xu, and Vladlen Koltun.
\newblock Learning to see in the dark.
\newblock In {\em 2018 IEEE/CVF Conference on Computer Vision and Pattern
  Recognition}, pages 3291--3300, June 2018.

\bibitem{mmdetection}
Kai Chen, Jiaqi Wang, Jiangmiao Pang, Yuhang Cao, Yu Xiong, Xiaoxiao Li,
  Shuyang Sun, Wansen Feng, Ziwei Liu, Jiarui Xu, Zheng Zhang, Dazhi Cheng,
  Chenchen Zhu, Tianheng Cheng, Qijie Zhao, Buyu Li, Xin Lu, Rui Zhu, Yue Wu,
  Jifeng Dai, Jingdong Wang, Jianping Shi, Wanli Ouyang, Chen~Change Loy, and
  Dahua Lin.
\newblock {MMDetection}: Open mmlab detection toolbox and benchmark.
\newblock {\em arXiv preprint arXiv:1906.07155}, 2019.

\bibitem{Zero-DCE}
Chunle Guo, Chongyi Li, Jichang Guo, Chen~Change Loy, Junhui Hou, Sam Kwong,
  and Cong Runmin.
\newblock Zero-reference deep curve estimation for low-light image enhancement.
\newblock {\em CVPR}, 2020.

\bibitem{LIME}
Xiaojie Guo, Yu Li, and Haibin Ling.
\newblock Lime: Low-light image enhancement via illumination map estimation.
\newblock {\em IEEE Transactions on Image Processing}, 26(2):982--993, 2017.

\bibitem{he2016deep}
Kaiming He, Xiangyu Zhang, Shaoqing Ren, and Jian Sun.
\newblock Deep residual learning for image recognition.
\newblock In {\em Proceedings of the IEEE conference on computer vision and
  pattern recognition}, pages 770--778, 2016.

\bibitem{Hong2021Crafting}
Yang Hong, Kaixuan Wei, Linwei Chen, and Fu Ying.
\newblock Crafting object detection in very low light.
\newblock In {\em BMVC}, 2021.

\bibitem{ignatov2020replacing}
Andrey Ignatov, Luc Van~Gool, and Radu Timofte.
\newblock Replacing mobile camera isp with a single deep learning model.
\newblock In {\em Proceedings of the IEEE/CVF Conference on Computer Vision and
  Pattern Recognition Workshops}, pages 536--537, 2020.

\bibitem{NIPS2015_33ceb07b}
Max Jaderberg, Karen Simonyan, Andrew Zisserman, and koray kavukcuoglu.
\newblock Spatial transformer networks.
\newblock In C. Cortes, N. Lawrence, D. Lee, M. Sugiyama, and R. Garnett,
  editors, {\em Advances in Neural Information Processing Systems}, volume~28.
  Curran Associates, Inc., 2015.

\bibitem{paa-eccv2020}
Kang Kim and Hee~Seok Lee.
\newblock Probabilistic anchor assignment with iou prediction for object
  detection.
\newblock In {\em ECCV}, 2020.

\bibitem{adam}
Diederik~P Kingma and Jimmy Ba.
\newblock Adam: A method for stochastic optimization.
\newblock {\em arXiv preprint arXiv:1412.6980}, 2014.

\bibitem{lamba2020towards}
Mohit Lamba, Atul Balaji, and Kaushik Mitra.
\newblock Towards fast and light-weight restoration of dark images.
\newblock {\em arXiv preprint arXiv:2011.14133}, 2020.

\bibitem{lamba2021restoring}
Mohit Lamba and Kaushik Mitra.
\newblock Restoring extremely dark images in real time.
\newblock In {\em Proceedings of the IEEE/CVF Conference on Computer Vision and
  Pattern Recognition}, pages 3487--3497, 2021.

\bibitem{REG}
Jinxiu Liang, Jingwen Wang, Yuhui Quan, Tianyi Chen, Jiaying Liu, Haibin Ling,
  and Yong Xu.
\newblock Recurrent exposure generation for low-light face detection.
\newblock {\em IEEE Transactions on Multimedia}, pages 1--1, 2021.

\bibitem{CameraNet}
Zhetong Liang, Jianrui Cai, Zisheng Cao, and Lei Zhang.
\newblock Cameranet: A two-stage framework for effective camera isp learning.
\newblock {\em IEEE Transactions on Image Processing}, 30:2248--2262, 2021.

\bibitem{lin2017focal}
Tsung-Yi Lin, Priya Goyal, Ross Girshick, Kaiming He, and Piotr Doll{\'a}r.
\newblock Focal loss for dense object detection.
\newblock In {\em Proceedings of the IEEE international conference on computer
  vision}, pages 2980--2988, 2017.

\bibitem{COCO}
Tsung-Yi Lin, Michael Maire, Serge Belongie, James Hays, Pietro Perona, Deva
  Ramanan, Piotr Doll{\'a}r, and C~Lawrence Zitnick.
\newblock Microsoft coco: Common objects in context.
\newblock In {\em European conference on computer vision}, pages 740--755.
  Springer, 2014.

\bibitem{liu2020joint}
Lin Liu, Xu Jia, Jianzhuang Liu, and Qi Tian.
\newblock Joint demosaicing and denoising with self guidance.
\newblock In {\em Proceedings of the IEEE/CVF Conference on Computer Vision and
  Pattern Recognition}, pages 2240--2249, 2020.

\bibitem{Exdark}
Yuen~Peng Loh and Chee~Seng Chan.
\newblock Getting to know low-light images with the exclusively dark dataset.
\newblock {\em Computer Vision and Image Understanding}, 178:30--42, 2019.

\bibitem{morawski2021nod}
Igor Morawski, Yu-An Chen, Yu-Sheng Lin, and Winston~H Hsu.
\newblock Nod: Taking a closer look at detection under extreme low-light
  conditions with night object detection dataset.
\newblock {\em arXiv preprint arXiv:2110.10364}, 2021.

\bibitem{omid2014pascalraw}
A Omid-Zohoor, D Ta, and B Murmann.
\newblock Pascalraw: raw image database for object detection, 2014.

\bibitem{paszke2019pytorch}
Adam Paszke, Sam Gross, Francisco Massa, Adam Lerer, James Bradbury, Gregory
  Chanan, Trevor Killeen, Zeming Lin, Natalia Gimelshein, Luca Antiga, et~al.
\newblock Pytorch: An imperative style, high-performance deep learning library.
\newblock {\em Advances in neural information processing systems}, 32, 2019.

\bibitem{qian2019rethinking}
Guocheng Qian, Yuanhao Wang, Chao Dong, Jimmy~S Ren, Wolfgang Heidrich, Bernard
  Ghanem, and Jinjin Gu.
\newblock Rethinking the pipeline of demosaicing, denoising and
  super-resolution.
\newblock {\em arXiv preprint arXiv:1905.02538}, 2019.

\bibitem{Ren_2017}
Shaoqing Ren, Kaiming He, Ross Girshick, and Jian Sun.
\newblock Faster r-cnn: Towards real-time object detection with region proposal
  networks.
\newblock {\em IEEE Transactions on Pattern Analysis and Machine Intelligence},
  Jun 2017.

\bibitem{rawpy}
Maik Riechert.
\newblock Raw image processing for python, a wrapper for libraw.
\newblock \url{https://github.com/letmaik/rawpy}, 2021.

\bibitem{sasagawa2020yolo}
Yukihiro Sasagawa and Hajime Nagahara.
\newblock Yolo in the dark-domain adaptation method for merging multiple
  models.
\newblock In {\em European Conference on Computer Vision}, pages 345--359.
  Springer, 2020.

\bibitem{schwartz2018deepisp}
Eli Schwartz, Raja Giryes, and Alex~M Bronstein.
\newblock Deepisp: Toward learning an end-to-end image processing pipeline.
\newblock {\em IEEE Transactions on Image Processing}, 28(2):912--923, 2018.

\bibitem{sharma2021nighttime}
Aashish Sharma and Robby~T Tan.
\newblock Nighttime visibility enhancement by increasing the dynamic range and
  suppression of light effects.
\newblock In {\em Proceedings of the IEEE/CVF Conference on Computer Vision and
  Pattern Recognition}, pages 11977--11986, 2021.

\bibitem{tian2019fcos}
Zhi Tian, Chunhua Shen, Hao Chen, and Tong He.
\newblock Fcos: Fully convolutional one-stage object detection.
\newblock {\em arXiv preprint arXiv:1904.01355}, 2019.

\bibitem{wang2021hla}
Wenjing Wang, Wenhan Yang, and Jiaying Liu.
\newblock Hla-face: Joint high-low adaptation for low light face detection.
\newblock In {\em 2021 IEEE/CVF Conference on Computer Vision and Pattern
  Recognition (CVPR)}, pages 16190--16199, 2021.

\bibitem{xing2021end}
Wenzhu Xing and Karen Egiazarian.
\newblock End-to-end learning for joint image demosaicing, denoising and
  super-resolution.
\newblock In {\em Proceedings of the IEEE/CVF Conference on Computer Vision and
  Pattern Recognition}, pages 3507--3516, 2021.

\bibitem{poor_visibility_benchmark}
Wenhan Yang, Ye Yuan, Wenqi Ren, Jiaying Liu, Walter~J. Scheirer, Zhangyang
  Wang, Taiheng Zhang, Qiaoyong Zhong, Di Xie, Shiliang Pu, Yuqiang Zheng,
  Yanyun Qu, Yuhong Xie, Liang Chen, Zhonghao Li, Chen Hong, Hao Jiang, Siyuan
  Yang, Yan Liu, Xiaochao Qu, Pengfei Wan, Shuai Zheng, Minhui Zhong, Taiyi Su,
  Lingzhi He, Yandong Guo, Yao Zhao, Zhenfeng Zhu, Jinxiu Liang, Jingwen Wang,
  Tianyi Chen, Yuhui Quan, Yong Xu, Bo Liu, Xin Liu, Qi Sun, Tingyu Lin,
  Xiaochuan Li, Feng Lu, Lin Gu, Shengdi Zhou, Cong Cao, Shifeng Zhang, Cheng
  Chi, Chubing Zhuang, Zhen Lei, Stan~Z. Li, Shizheng Wang, Ruizhe Liu, Dong
  Yi, Zheming Zuo, Jianning Chi, Huan Wang, Kai Wang, Yixiu Liu, Xingyu Gao,
  Zhenyu Chen, Chang Guo, Yongzhou Li, Huicai Zhong, Jing Huang, Heng Guo,
  Jianfei Yang, Wenjuan Liao, Jiangang Yang, Liguo Zhou, Mingyue Feng, and
  Likun Qin.
\newblock Advancing image understanding in poor visibility environments: A
  collective benchmark study.
\newblock {\em IEEE Transactions on Image Processing}, 29:5737--5752, 2020.

\end{thebibliography}
}

\end{document}